\documentclass{article}
\PassOptionsToPackage{numbers,compress}{natbib}
\usepackage[preprint]{neurips_2026}
\usepackage[table]{xcolor}
\usepackage{booktabs}
\usepackage{array}
\usepackage{marvosym}
\usepackage[utf8]{inputenc}
\usepackage[T1]{fontenc}
\usepackage{hyperref}
\usepackage{url}
\usepackage{amsfonts}
\usepackage{amsmath}
\usepackage{amssymb}
\usepackage{nicefrac}
\usepackage{microtype}
\usepackage{graphicx}
\usepackage{array}
\usepackage{multirow}
\usepackage{enumitem}
\usepackage{tikz}
\usepackage{tcolorbox}
\tcbuselibrary{skins,breakable,listingsutf8}
\usetikzlibrary{arrows.meta,positioning,fit,calc,backgrounds,shapes.geometric}

\title{GenEvolve: Self-Evolving Image Generation Agents via Tool-Orchestrated Visual Experience Distillation}

\author{
Sixiang Chen$^{1,2}$\
Zhaohu Xing$^{1}$\
Tian Ye$^{1}$\
Xinyu Geng$^{3}$\
Yunlong Lin \
\textbf{Jianyu Lai$^{1,2}$} \\
\textbf{Xuanhua He$^{3}$}\
\textbf{Fuxiang Zhai$^{1}$}\
\textbf{Jialin Gao$^{4,\dag}$}\
\textbf{Lei Zhu$^{1,3}$}\textsuperscript{\Letter} \\
$^{1}$The Hong Kong University of Science and Technology (Guangzhou) 
$^{2}$Meituan \\
$^{3}$The Hong Kong University of Science and Technology \\
$^{4}$National University of Singapore \\
Project Repo: \url{https://ephemeral182.github.io/GenEvolve/}
}

\newcommand{\method}{GenEvolve}
\newcommand{\ved}{Visual Experience Distillation}
\newcommand{\tool}[1]{\texttt{#1}}
\newcommand{\slot}[1]{\textsc{#1}}
\newcommand{\skill}[1]{\textit{#1}}
\definecolor{GEInk}{HTML}{111827}
\definecolor{GEMuted}{HTML}{6B7280}
\definecolor{GEBlue}{HTML}{2563EB}
\definecolor{GECyan}{HTML}{0891B2}
\definecolor{GEGreen}{HTML}{16A34A}
\definecolor{GEAmber}{HTML}{D97706}
\definecolor{GERed}{HTML}{DC2626}
\definecolor{GELight}{HTML}{F3F4F6}
\definecolor{GEBlueLight}{HTML}{EFF6FF}
\definecolor{GECyanLight}{HTML}{ECFEFF}
\definecolor{GEGreenLight}{HTML}{ECFDF5}
\definecolor{GEAmberLight}{HTML}{FFFBEB}
\definecolor{GERedLight}{HTML}{FEF2F2}

\newcommand{\smallpill}[2]{\begingroup\setlength{\fboxsep}{1.6pt}\colorbox{#1!12}{\textcolor{#1!80!black}{\scriptsize #2}}\endgroup}
\tcbset{
  examplebox/.style={
    colback=green!5!white,
    colframe=green!75!black,
    coltitle=green!50!black,
    fonttitle=\bfseries,
    boxrule=0.5mm,
    sharp corners,
    enhanced,
    left=2mm,
    right=2mm,
    top=2mm,
    bottom=2mm,
    attach boxed title to top left={
      yshift=-2mm,
      xshift=5mm
    },
    boxed title style={
      colframe=green!75!black,
      colback=white,
      sharp corners
    }
  }
}
\newtcolorbox[auto counter, number within=section]{example}[2][]{%
  examplebox,
  title=Prompt~\thetcbcounter~(#2),
  #1
}
\newtcblisting{promptlisting}[2][]{%
  examplebox,
  breakable,
  listing only,
  title=Prompt~(#2),
  listing options={
    basicstyle=\ttfamily\small,
    breaklines=true,
    breakatwhitespace=false,
    columns=fullflexible,
    keepspaces=true,
    showstringspaces=false,
    tabsize=2
  },
  #1
}

\begin{document}

\maketitle

\begin{abstract}
Open-ended image generation is no longer a simple prompt-to-image problem. High-quality generation often requires an agent to combine a model's internal generative ability with external resources. As requests become more diverse and demanding, we aim to develop a general image-generation agent that can self-evolve through trajectories and use tools more effectively across varied generation challenges.
To this end, we propose \textbf{\method{}}, a self-evolving framework based on Tool-Orchestrated Visual Experience Distillation. In \method{}, each generation attempt is modeled as a tool-orchestrated trajectory, where the agent gathers evidence, selects references, invokes generation skills, and composes them into a prompt-reference program. Unlike existing agentic generation methods that mainly rely on image-level scalar rewards, \method{} compares multiple trajectories for the same request and abstracts best-worst differences into structured visual experience, provided only to a privileged teacher branch. Inspired by on-policy self-distillation, \ved{} provides dense token-level supervision, helping the student internalize better search, knowledge activation, reference selection, and prompt construction.
We further construct GenEvolve-Data and GenEvolve-Bench. Experiments on public benchmarks and GenEvolve-Bench show substantial gains over strong baselines, achieving state-of-the-art performance among current image-generation frameworks.
\end{abstract}

% \vspace{-0.3cm}
\section{Introduction}
% \vspace{-0.2cm}

\begin{figure*}[!t]
    \centering
    \includegraphics[width=\textwidth]{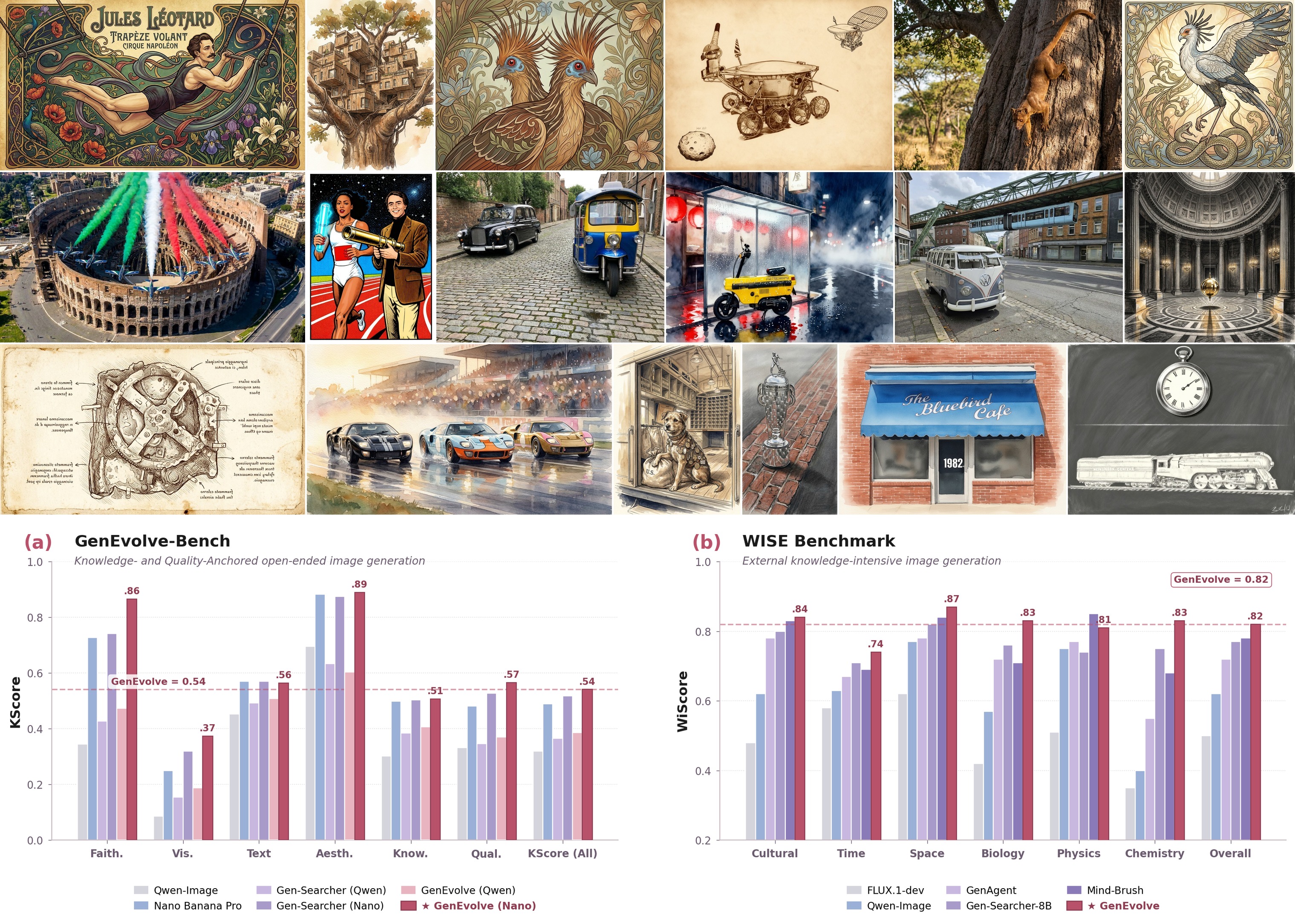}
    \caption{
\textbf{Results of \method{}.}
\textit{Top:} Representative generation results by our self-evolving agent across diverse open-ended and complicated requests covering architecture, creative transfer, scientific illustration, street scenes, and more, using both Nano Banana Pro and Qwen-Image-Edit as downstream generators.
\textit{Bottom:} Quantitative comparison on \textbf{(a)} our \textsc{GenEvolve-Bench} (KScore + four judge dimensions and Knowledge-/Quality-Anchored tracks) and \textbf{(b)} the external \textsc{WISE} benchmark, where \method{} consistently outperforms SOTA direct generators and recent agentic baselines.
}
    \label{fig:teaser}
    \vspace{-4mm}
\end{figure*}

Modern image generators are increasingly powerful, but open-ended image generation is not solved by fidelity alone. Real requests require deciding what the generator already knows, what external facts and references to acquire, which internal generation knowledge to activate, and how to translate these signals into instructions a downstream generator can follow. Thus, high-quality generation is becoming less a one-shot prompt-to-image task than an agentic process of planning, tool orchestration, and feedback-driven adaptation.

This shift is most visible in complex and grounded generation scenarios. 
A request may involve current or long-tail factual knowledge, reference-specific appearance, multi-source visual evidence, professional design constraints, or implicit user intent that cannot be captured by a single rewritten prompt. 
Strong generators may possess substantial internal knowledge and visual priors, but they do not decide when to search, how to use internal knowledge, which references are useful, or how failures should guide future behavior. 
Thus, the key challenge is not simply improving local abilities such as text rendering, layout, counting, or attribute binding. 
Rather, it is to build a general image-generation agent that can coordinate internal generative knowledge with external tools and learn how to use them through interaction with the generator. Such coordination requires more than exposing a tool list: the agent must learn when a request needs factual lookup, what queries should be issued, which retrieved images should serve as references, which generation knowledge should be activated via skills, and how these signals should be bound into a generator-facing program.

Recent agentic generation systems have begun to explore this direction. 
GenAgent treats image generators as invokable tools for multi-turn reasoning, tool use, judgment, and reflection~\citep{jiang2026genagent}. 
Gen-Searcher and ORIG improve factual grounding through search- or retrieval-augmented generation~\citep{gensearcher,orig}, while GEMS and Mind-Brush introduce memory, reusable skills, or research-style workflows~\citep{gems,he2026mind}. 
Maestro and CRAFT further refine generation with critic feedback, verifier agents, or constraint-driven correction~\citep{maestro,craft}. 
These systems show the value of search, tools, memory, and iterative refinement, but they usually address only part of the generation process: acquiring external evidence, wrapping a black-box generator, or evolving prompts at inference time. 
It therefore remains underexplored how to train an open image-generation agent whose tool use, reference selection, knowledge activation, prompt-reference program construction, and generator interaction are optimized together.

Therefore, we propose \method{}, a self-evolving framework for image-generation agents based on \textbf{Tool-Orchestrated Visual Experience Distillation}. 
\method{} models each generation attempt as a tool-orchestrated visual trajectory, in which the agent gathers textual evidence, retrieves and selects visual references, invokes callable generation knowledge, and synthesizes a prompt-reference program \(z=(g,R)\), where \(g\) is a targeted prompt and \(R\) is a small set of selected reference images. 
A reference-conditioned generator then produces the final image, which is evaluated together with the trajectory that produced it via reward calculation and diagnostics. 
Thus, the learning target is not merely a prompt, but a complete generation trajectory linking tool decisions, generator-facing instructions, generated outcomes, and feedback.

To make this formulation trainable and measurable, we construct \textbf{GenEvolve-Data} and \textbf{GenEvolve-Bench}. 
GenEvolve-Data goes beyond ordinary prompt-rewriting corpora by providing tool-orchestrated trajectories that teach the agent how to acquire external evidence, activate internal generation knowledge, and construct prompt-reference programs. It further provides filtered GT image cases that make visual feedback meaningful for self-evolution. 
GenEvolve-Bench evaluates final image quality across Knowledge-Anchored and Quality-Anchored settings, covering both external grounding and quality-sensitive generation requirements. 

On top of this trajectory data, \method{} turns visual outcomes into structured experience for improving the agent. Existing agentic generation methods can optimize trajectories with image-level scalar rewards, but such rewards indicate which trajectory is better without explaining which decisions caused the improvement. \method{} instead compares multiple trajectories for the same request and abstracts best-worst differences into visual experience. Inspired by on-policy self-distillation, this experience is provided only to a privileged teacher branch, while the student acts under the normal inference context. Combined with group-relative policy optimization, Visual Experience Distillation provides dense token-level supervision for better tool orchestration and generator-facing program synthesis. As illustrated in Figure~\ref{fig:teaser}, \method{} produces high-quality images across diverse open-ended requests and consistently outperforms strong direct generators and recent agentic baselines on both our \textsc{GenEvolve-Bench} and the external \textsc{WISE} benchmark. Therefore, our contributions are summarized as follow:
\begin{itemize}
\item We propose \textbf{GenEvolve}, which reformulates open-ended image generation as an agentic trajectory learning problem, where a general image-generation agent learns to coordinate internal generative knowledge with external tools, including factual search, visual reference retrieval, callable generation knowledge, prompt-reference program synthesis, image generation, and experience internalization.
\item We first introduce a self-evolving post-training mechanism that compares multiple trajectories for the same request and abstracts best-worst trajectory differences into structured visual experience. The token-level distillation objective builds on established on-policy self-distillation losses, while our contribution is the visual experience construction, retrieval, and teacher-only conditioning for image-generation agents.
\item We construct a trajectory dataset and diagnostic benchmark for general image-generation agents, evaluating both final image quality and agentic behaviors such as tool use, reference selection, skill routing, and prompt-reference faithfulness.
\item Experiments show that \method{} achieves the best performance on GenEvolve-Bench and the public benchmark, outperforming raw generators, agentic baselines and further improving with a stronger generator. These results demonstrate the effectiveness and transferability of the learned prompt-reference programs and tool-orchestrated policy.

\end{itemize}

% \vspace{-0.1cm}
\section{Related Work}
% \vspace{-0.2cm}
\noindent\textbf{Image generation models.}
Image generation has evolved from standalone text-to-image generators to integrated multimodal generation systems. Diffusion and latent diffusion models established high-fidelity prompt-conditioned synthesis~\citep{rombach2022stable,saharia2022imagen,ramesh2022hierarchical,chen2025postercraft,chen2026posteromni}, while diffusion transformers and their successors, including DiT, PixArt-$\alpha$, Stable Diffusion 3, FLUX, Hunyuan-DiT, and Nano Banana Pro, further improve scalability, text understanding, and generation quality~\citep{peebles2022dit,chen2023pixartalpha,esser2024sd3,flux2024,hunyuan2024dit,google2025nanobananapro}. In parallel, unified multimodal models such as Chameleon, Emu3, Show-o, BAGEL, OmniGen2, HunyuanImage 3.0, and BLIP3-o explore shared or hybrid architectures for multimodal understanding and generation~\citep{chameleon2024,wang2024emu3,xie2024showo,bagel2025,wu2025omnigen2,cao2025hunyuanimage3,chen2025blip3o}. Despite their strong rendering ability and multimodal flexibility, these models remain primarily generators: they do not explicitly decide when to acquire missing facts, which references to trust, or which generation knowledge to activate.

\noindent\textbf{Agentic image generation.}
Agentic generation systems augment image models with planning, retrieval, tool use, judging, or refinement. GenAgent enables multi-turn reasoning, tool invocation, judgment, and reflection around image generators~\citep{jiang2026genagent}. Mind-Brush~\citep{he2026mind}, Gen-Searcher, and ORIG focus on research/search/retrieval-augmented generation for implicit, dynamic, or factual knowledge~\citep{he2026mind,gensearcher,orig}. GEMS introduces memory and skills~\citep{gems}, while Maestro and CRAFT use critic/verifier feedback or constraint-driven correction to iteratively improve prompts~\citep{maestro,craft}. These systems show the value of search, tools, memory, and refinement, but they often emphasize one component of the broader generation process or wrap a generator with an external workflow. Recent commercial systems such as Nano Banana Pro, built on Gemini, point toward tighter integration of reasoning, real-world knowledge, grounding, and visual synthesis~\citep{google2025nanobananapro}. Inspired by this direction, \method{} trains an open image-generation agent that coordinates external tools and internal generation knowledge along visual trajectories, and uses visual experience distillation to improve the coupling between the agent policy and downstream generator behavior.

\noindent\textbf{On-policy distillation.}
On-policy distillation has become a promising post-training paradigm for language models and agents, with variants including OPSD, OPCD, Skill-SD, SDPO, and HDPO~\citep{opsd,opcd,skillsd,sdpo,hdpo}. OPSD uses privileged context to supervise on-policy generations~\citep{opsd}; OPCD distills useful in-context knowledge into model parameters~\citep{opcd}; SDPO converts rich feedback into dense self-distillation signals~\citep{sdpo}; and Skill-SD summarizes multi-turn agent trajectories into training-only skills with an importance-weighted sampled-token reverse-KL objective~\citep{skillsd}. \method{} is motivated by this general teacher-only self-distillation principle, but changes the privileged signal and task character: instead of ground-truth reasoning traces or text-agent skills, the teacher receives visual experience extracted from tool-orchestrated image-generation trajectories, helping the student internalize better search, knowledge activation, reference selection, and prompt-reference program synthesis.

% \vspace{-0.1cm}
\section{Tool-Orchestrated Visual Trajectory Formulation}
% \vspace{-0.1cm}

We formalize each generation attempt as a \emph{tool-orchestrated visual trajectory}. 
Given a user request \(x\), the agent does not directly generate an image, merely rewrite the prompt, or only retrieve external evidence. 
Instead, it decides when to acquire external information, which visual references to trust, when to activate internal generation knowledge, and how to synthesize these signals into a prompt-reference program. 
This makes the generation process observable and trainable, covering both external tool use and internal knowledge activation before generation.

At turn \(t\), the agent observes the interaction history and samples an action:
\begin{equation}
\begin{aligned}
h_t &= (x,a_1,o_1,\ldots,a_{t-1},o_{t-1}),\\
a_t &\sim \pi_\theta(a\mid h_t), \qquad
o_t \sim \mathcal{T}_{a_t}(o\mid h_t),
\end{aligned}
\end{equation}
where \(a_t\) is either a tool call or the final answer, and \(o_t\) is the corresponding observation. 
The final answer is a prompt-reference generation program \(z=(g,R)\), where \(g\) is a targeted generation prompt and \(R\) is a small set of selected reference images. 
A reference-conditioned generator renders \(\hat{y}=G(g,R)\). 
A complete trajectory is therefore
\begin{equation}
\tau=(x,a_1,o_1,\ldots,a_T,o_T,z,\hat{y},r,d),
\end{equation}
where \(r\) is a scalar reward and \(d\) contains visual diagnostics. 
The trajectory-level objective is
\begin{equation}
\max_\theta \; \mathbb{E}_{x,\tau}\!\left[R(\hat{y},z,x)\right].
\end{equation}

In \method{}, however, reward is not the only learning signal. 
For the same request, multiple trajectories may produce different visual outcomes. 
\method{} compares high- and low-reward trajectories and converts their differences into structured visual experience \(M\), which is provided only to a privileged teacher branch during self-distillation.

This formulation differs from prior agentic generation systems in both optimization scope and supervision source. 
Many existing methods expose external interfaces such as search, retrieval, judging, or prompt correction around black-box or loosely coupled generators. 
\method{} instead treats the whole generation process as the learnable object: external tool use, internal generation-knowledge activation, reference selection, and prompt-reference synthesis are modeled as trajectory decisions. 
By distilling visual experience into the student policy, \method{} teaches not only which trajectory is better, but which orchestration behaviors should be reused for future requests.

% \vspace{-0.2cm}
\section{GenEvolve-Data and GenEvolve-Bench}
% \vspace{-0.2cm}
\label{sec:data_bench}

\begin{figure*}[!t]
    \centering
    \includegraphics[width=14cm]{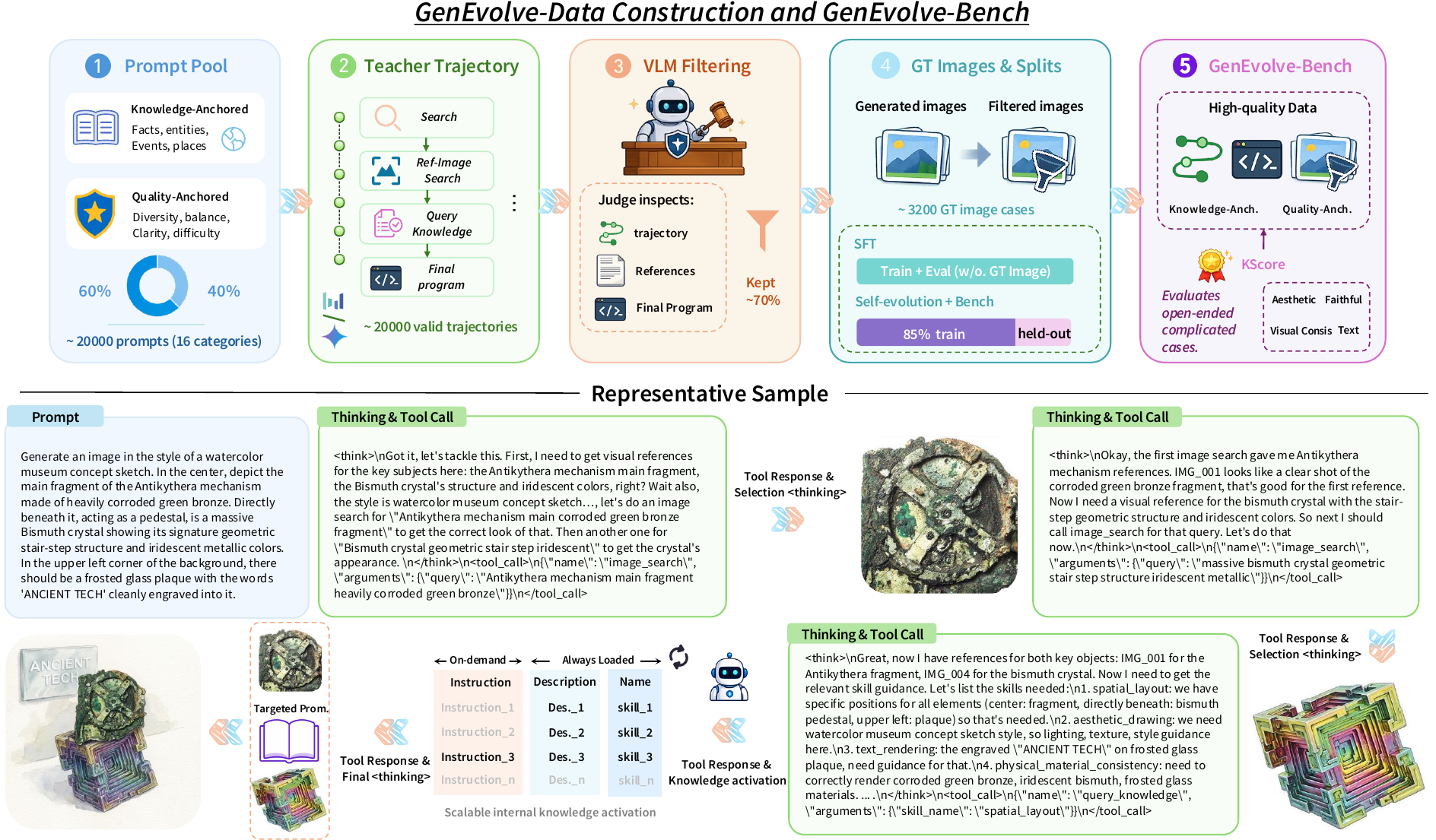}
   \caption{
\textbf{Overview of GenEvolve-Data and GenEvolve-Bench.}
The top row presents the construction pipeline: diverse prompts are converted into tool-orchestrated teacher trajectories, audited by VLM-based checks, used to generate and filter GT image cases, and split for supervised training, self-evolution, and held-out evaluation.
The bottom row illustrates a representative case, showing how the agent retrieves visual references, activates generation knowledge, and composes a prompt-reference program for grounded image generation.
}
    \label{fig:data_pipeline}
    \vspace{-4mm}
\end{figure*}

Before introducing the learning algorithm, we first define the data substrate that enables tool-orchestrated visual trajectory learning. As shown in Figure~\ref{fig:data_pipeline}, GenEvolve-Data is constructed as a complete generation pipeline rather than a prompt-rewriting corpus: diverse prompts are solved by teacher agents through tool use, audited by VLM filters, rendered into GT image cases, and split for supervised cold start, self-evolution, and held-out evaluation.

\noindent\textbf{Prompt pool.}
We construct natural user requests from structured recipes specifying the task family, missing external evidence, visual anchor, dominant generation requirement, and difficulty. The pool contains two complementary tracks. \emph{Knowledge-Anchored} prompts require external grounding for entities, events, places, objects, or visual facts, while \emph{Quality-Anchored} prompts emphasize quality-sensitive generation requirements such as text layout, spatial composition, counting, anatomy, material consistency, aesthetics, and creative transfer. These recipe fields are used for coverage control and stratified splitting, but are not exposed to the agent as task labels.

\noindent\textbf{Teacher trajectories.}
Each validated prompt is converted into a teacher trajectory through a real multi-turn tool loop. We use Seed2.0 and Gemini 3 Pro as teacher models, leveraging their multimodal understanding, reasoning, and agentic capabilities to issue textual search queries, retrieve visual references, activate generation knowledge, and synthesize the final prompt-reference program~\citep{seed2026seed2,google2025gemini3pro}. The tool order is request-dependent: knowledge-heavy cases may begin with factual lookup, reference-sensitive cases rely more on image search, and quality-anchored cases may activate generation knowledge for text, layout, pose, material, or style control. Each trajectory records the tool observations, selected references, intermediate rationale, and final program used for generation.

\noindent\textbf{Trajectory filtering.}
We audit teacher trajectories before using them for training. Programmatic checks remove incomplete tool loops, invalid reference selections, raw URL or ID leakage, missing ordinal reference wording, and underspecified final programs. A VLM judge then reviews whether the selected references support the requested visual details, whether the collected evidence is actually used, and whether the final program integrates the required constraints. This produces a high-quality trajectory set for SFT cold start.

\noindent\textbf{GT images and splits.}
For self-evolution and evaluation, high-quality teacher programs are rendered into GT image cases using Nano Banana Pro, which is built on Gemini 3 Pro Image and is designed for high-quality image generation/editing with strong text rendering, visual control, and real-world knowledge~\citep{google2025nanobananapro}. A second visual filter checks prompt compliance, reference usage, visual coherence, and image quality. The surviving cases are exported into two views: an SFT view that preserves full tool-loop trajectories without exposing GT images, and a visual-feedback view that contains the user request, GT image, and metadata for self-evolution and benchmark evaluation. This design enables supervised cold start while preventing the self-evolving agent from simply copying teacher outputs.

\noindent\textbf{GenEvolve-Bench.}
GenEvolve-Bench is the held-out evaluation split produced by the same pipeline. It evaluates open-ended image-generation agents under a unified KScore~\citep{gensearcher} protocol, with results reported on both Knowledge-Anchored and Quality-Anchored subsets. The benchmark is designed to test whether agents can combine external evidence, selected visual references, and quality-aware generation control, rather than merely follow generic text-to-image prompts. Overall details and construction statistics of our data are provided in Appendix~\ref{sec:appendix_data_construction}.

% \vspace{-0.2cm}
\section{Method: \method}
% \vspace{-0.2cm}
\label{sec:method}

\begin{figure*}[!t]
    \centering
    \includegraphics[width=14cm]{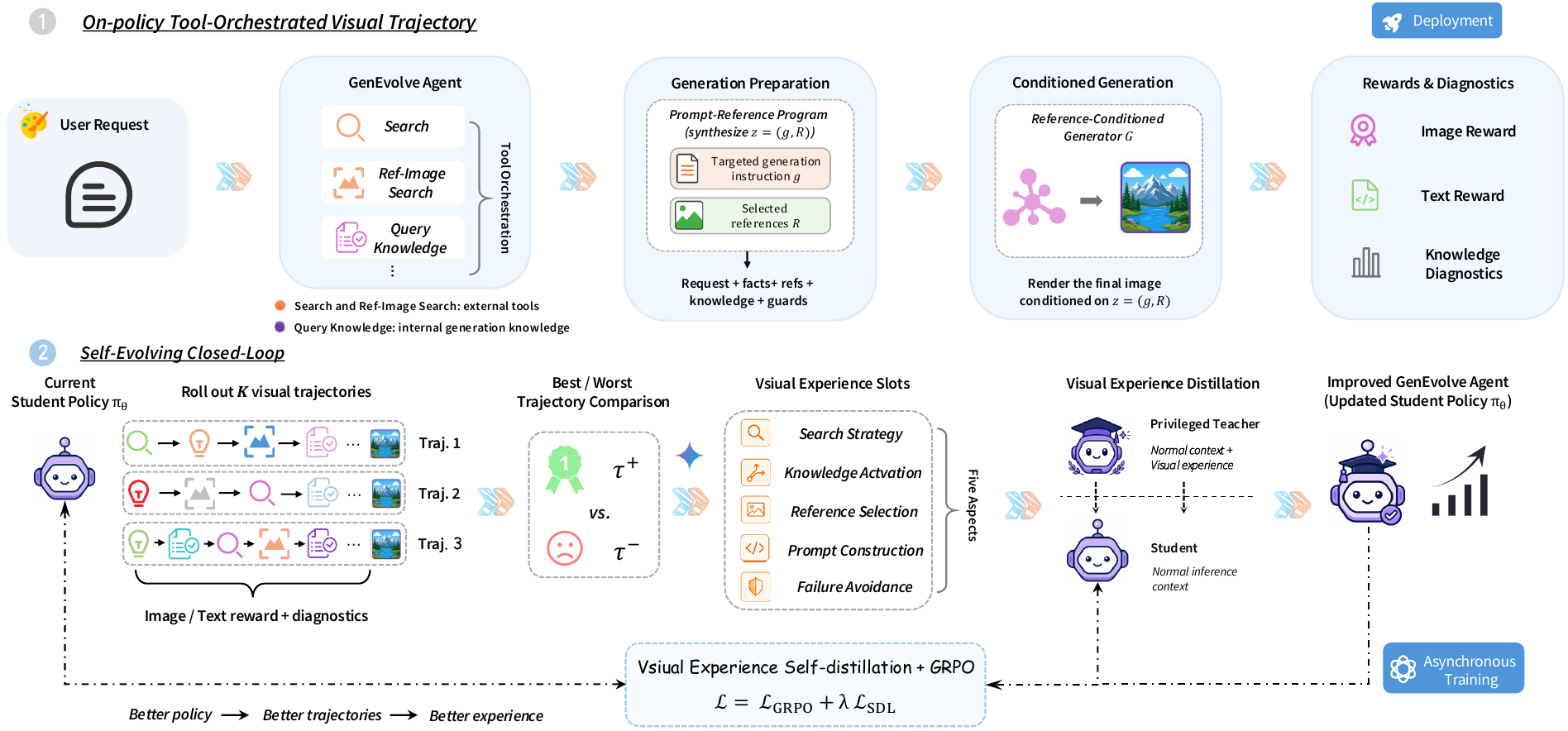}
\caption{
\textbf{Overview of \method{}.}
The student agent orchestrates external search, visual references, and internal generation knowledge to produce a prompt-reference program \(z=(g,R)\).
During training, multiple trajectories are judged with image/text rewards; best-worst differences are converted into visual experience and injected only into a privileged teacher.
GRPO provides trajectory-level optimization, while Visual Experience Self-Distillation supplies token-level guidance, forming a self-evolving loop of better policy, trajectories, and experience.
}
    \label{fig:overview}
    \vspace{-4mm}
\end{figure*}

\subsection{Tool-Orchestrated Visual Trajectories}

Given the trajectory data and GT image cases in Section~\ref{sec:data_bench}, \method{} trains an image-generation agent whose output is produced through a multi-turn visual trajectory rather than a single prompt rewrite. 
For a user request \(x\), the agent samples \(\tau=(a_1,o_1,\ldots,a_T,o_T,z)\), where each \(a_t\) is a tool call or the final answer, \(o_t\) is the corresponding observation, and \(z\) is an executable prompt-reference generation program. 
Following a ReAct-style interface, search planning, reference acquisition, internal-knowledge activation, and prompt-reference program construction become explicit trajectory decisions.

The action space contains three tool families. 
\tool{search(q)} gathers textual evidence for visible facts, \tool{image\_search(q)} retrieves candidate visual references, and \tool{query\_knowledge(skill\_name)} activates internal generation knowledge. 
We instantiate such internal knowledge as compact callable generation skills, covering text rendering, layout, counting, anatomy, attribute binding, material consistency, aesthetics, and creative transformation. 
Static generation knowledge remains available to the deployed student through tool calls, while dynamic visual experience is used only during training. 
The full tool protocol and skill taxonomy are provided in the appendix.

\subsection{SFT Cold Start for Tool-Orchestrated Agents}

GenEvolve-Data first provides supervised trajectories to cold-start the base MLLM into a tool-orchestrated image-generation agent. 
This stage teaches the model to follow the visual trajectory formulation: when to call tools, how to retrieve and select references, when to activate internal generation knowledge, and how to output a valid prompt-reference program \(z=(g,R)\).

Each SFT example contains a user request, a multi-turn tool trajectory, and a final program:
\begin{equation}
\tau^{\star}=(a_1,o_1,\ldots,a_T,o_T,z^{\star}), 
\qquad 
z^{\star}=(g^{\star},R^{\star}).
\end{equation}
We optimize assistant-side trajectory tokens under the observed tool history:
\begin{equation}
\mathcal{L}_{\mathrm{SFT}}(\theta)
=
-\frac{1}{\sum_{i,t}m_{i,t}}
\sum_i \sum_t
m_{i,t}
\log \pi_{\theta}(y_{i,t}^{\star}\mid h_{i,t}^{\star}),
\end{equation}
where \(h_{i,t}^{\star}\) includes previous tool observations and \(m_{i,t}\) masks valid assistant tokens. 
After this cold start, GRPO and Visual Experience Distillation further optimize the initialized policy with generated-image feedback, forming the self-evolving stage shown in Fig.~\ref{fig:overview}.

\subsection{Prompt-Reference Program and Generation Feedback}

The final trajectory output is a prompt-reference generation program
\begin{equation}
  z=(g,R), \qquad R=\{r_1,\ldots,r_k\},
\end{equation}
where \(g\) is a targeted generation instruction and \(R\) is an ordered set of selected reference images. 
The instruction refers to selected images by ordinal phrases such as ``the first reference image'', rather than raw URLs or retrieval IDs. 
Program synthesis binds constraints from the user request, retrieved facts, selected references, activated internal generation knowledge, and failure-avoidance experience:
\begin{equation}
  C=C_{\mathrm{user}}\cup C_{\mathrm{fact}}\cup C_{\mathrm{ref}}
    \cup C_{\mathrm{know}}\cup C_{\mathrm{avoid}}.
\end{equation}
A reference-conditioned generator then produces
\begin{equation}
  \hat{y}=G(g,R).
\end{equation}

For trajectory-level optimization, we follow recent work Gen-Searcher~\citep{gensearcher} and use dual reward feedback: an image-side reward evaluates the generated image, while a text-side reward evaluates the agent's final program. 
Specifically, \(R_{\mathrm{img}}\) follows the KScore-style image judge over faithfulness, visual correctness, text accuracy, and aesthetics. 
Different from a generic fluency or prompt-quality score, our \(R_{\mathrm{text}}\) is designed as a \emph{program sufficiency reward}: it checks whether \(z=(g,R)\) contains enough grounded facts, ordinal reference bindings, activated generation knowledge, and executable generation constraints for a strong generator to reproduce the intended image. 
The final reward is
\begin{equation}
  R=(1-\alpha)R_{\mathrm{img}}+\alpha R_{\mathrm{text}}.
\end{equation}

For each request, \method{} samples \(K\) trajectories and applies standard group-relative policy optimization over the mixed reward:
\begin{equation}
  \widehat{A}_i=
  \frac{R_i-\bar{R}}{\sigma_R+\epsilon_{\mathrm{adv}}}.
\end{equation}
We optimize the clipped GRPO surrogate over assistant-side trajectory tokens, following prior GRPO-based policy optimization~\citep{deepseekmath,gensearcher,qwen3vl}. 
This reward-driven term identifies which visual trajectories are better, while the visual experience distillation term in Section~\ref{subsec:visual_experience_distillation} provides denser token-level guidance about why the better trajectory should be preferred. 
The full GRPO objective is given in Appendix~\ref{subsec:grpo_loss}, and the concrete training hyper-parameters (clip ratios, KL coefficient, batch sizes, etc.) are summarised in Table~\ref{tab:appendix_rl}.

\vspace{-2mm}
\subsection{Visual Experience Extraction}

Scalar rewards indicate which trajectories are better, but not why they are better. 
\method{} therefore converts generated outcomes into structured visual experience. 
For the \(K\) trajectories sampled for the same request, we identify the best and worst trajectories:
\begin{equation}
  \tau^+=\arg\max_i R(\tau_i),\qquad
  \tau^-=\arg\min_i R(\tau_i),\qquad
  \Delta=R(\tau^+)-R(\tau^-).
\end{equation}
If \(\Delta\ge\delta_{\min}\), the best-worst pair is summarized into five experience slots:
\begin{equation}
M=\{
M_{\mathrm{search}},
M_{\mathrm{know}},
M_{\mathrm{ref}},
M_{\mathrm{prompt}},
M_{\mathrm{fail}}
\}.
\end{equation}
These slots capture the search strategy, internal-knowledge activation, reference selection, prompt-reference construction, and failure-avoidance lesson that distinguish the stronger trajectory from the weaker one. 
They are extracted by a VLM judge from the complete tool trajectories, final programs, judge rationales and corresponding diagnostics. 
Pairs caused only by protocol failures, such as missing references, are discarded because they do not provide reusable visual strategy.

The resulting experience buffer is prompt-keyed and used only during training. 
Each entry is attached to the source prompt that produced it and stores the corresponding prompt embedding, computed with Qwen3-Embedding-0.6B. 
For a new prompt \(x\), the teacher retrieves experience by source-prompt similarity rather than by matching the compact experience text:
\begin{equation}
  \tilde{x}=\arg\max_{x_j\in\mathcal{B}}\cos(e(x),e(x_j)),
  \qquad
  M_x=\{M_s(x_j):s\in\mathcal{S},x_j=\tilde{x}\}.
\end{equation}
This source-prompt bundle retrieval returns a coherent set of lessons from one historical case and avoids mixing unrelated slot entries. 
The detailed slot schema, memory update rule, and fallback strategy are provided in Appendix~\ref{subsec:source_prompt_retrieval}.

\begin{table}[t]
\centering
\footnotesize
\caption{Main comparison on GenEvolve-Bench. KScore follows the Gen-Searcher~\cite{gensearcher} visual judge. Know.-Anch. and Qual.-Anch. denotes the Knowledge-Anchored and Quality-Anchored tracks. Best results are in bold; underlined values mark the best open-generator agent result.}
\label{tab:main_results}
\setlength{\tabcolsep}{4pt}
\renewcommand{\arraystretch}{1.12}
\resizebox{\linewidth}{!}{%
\begin{tabular}{llccccccc}
\toprule
\rowcolor{GEBlue!10}
 & &
\multicolumn{4}{c}{Judge dimensions} &
\multicolumn{3}{c}{Benchmark overall} \\
\rowcolor{GEBlue!10}
\multirow{-2}{*}{\centering Method} &
\multirow{-2}{*}{\centering Generator} &
Faith. & Vis. & Text & Aesth. &
    KScore (All) & Know.-Anch. & Qual.-Anch. \\
\midrule

\multicolumn{9}{l}{\textit{Direct generator baselines \smallpill{GEMuted}{raw}}} \\
 Lumina-Image 2.0~\cite{qin2025luminaimage2}
& Lumina-Image 2.0
& 0.1044 & 0.0000 & 0.3308 & 0.2694
& 0.1697 & 0.1528 & 0.1915 \\

BAGEL~\cite{bagel2025}
& BAGEL
& 0.1212 & 0.0059 & 0.3721 & 0.4082
& 0.2041 & 0.1684 & 0.2504 \\

SD-3.5-Large~\cite{stabilityai2024sd35large}
& SD-3.5
& 0.1456 & 0.0135 & 0.3872 & 0.4865
& 0.2235 & 0.1943 & 0.2612 \\

FLUX.1-dev~\cite{flux2024}
& FLUX.1
& 0.1574 & 0.0059 & 0.4150 & 0.5556
& 0.2396 & 0.2097 & 0.2784 \\

FLUX.2 Klein 4B~\cite{flux2klein}
& FLUX.2
& 0.2525 & 0.0059 & 0.3847 & 0.5648
& 0.2380 & 0.2004 & 0.2865 \\

Z-Image-Turbo~\cite{cai2025z}
& Z-Image
& 0.2837 & 0.0396 & 0.4369 & 0.6187
& 0.2808 & 0.2340 & 0.3413 \\

FLUX.2 Klein 9B~\cite{flux2klein}
& FLUX.2
& 0.3662 & 0.0210 & 0.4192 & 0.6599
& 0.2787 & 0.2327 & 0.3382 \\

Z-Image~\cite{cai2025z}
& Z-Image
& 0.3333 & 0.0278 & 0.4352 & 0.5429
& 0.2728 & 0.2203 & 0.3407 \\

Qwen-Image~\cite{qwenimage}
& Qwen-Image
& 0.3729 & 0.0623 & 0.4226 & 0.6751
& 0.2987 & 0.2384 & 0.3768 \\

Nano Banana Pro~\cite{google2025nanobananapro}
& Nano Banana Pro
& 0.7761 & 0.2837 & 0.6178 & 0.9158
& 0.5298 & 0.5160 & 0.5477 \\

\midrule
\multicolumn{9}{l}{\textit{Agentic image-generation workflows \smallpill{GEBlue}{agent}}} \\
Gen-Searcher 8B~\cite{gensearcher}
& Qwen-Image-Edit-2511
& 0.5284 & 0.1050 & 0.4768 & \underline{0.6377}
& 0.3493 & 0.3293 & 0.3745 \\

Gen-Searcher 8B~\cite{gensearcher}
& Nano Banana Pro
& 0.7465 & 0.3378 & 0.6198 & 0.9036
& 0.5481 & 0.5472 & 0.5492 \\

% Qwen3-VL-Workflow
% & Qwen-Image-Edit-2511
% & 0.4781 & 0.1061 & 0.4571 & 0.5867
% & 0.3317 & 0.3109 & 0.3587 \\

\rowcolor{GELight!45}
\method{} (open generator)
& Qwen-Image-Edit-2511
& \underline{0.5303} & \underline{0.1338} & \underline{0.4907} & 0.6347
& \underline{0.3663} & \underline{0.3410} & \underline{0.3990} \\

\rowcolor{GEGreen!10}
\method{} (strong generator)
& Nano Banana Pro
& \textbf{0.7970} & \textbf{0.3832} & \textbf{0.6218} & \textbf{0.9222}
& \textbf{0.5739} & \textbf{0.5669} & \textbf{0.5830} \\
\bottomrule
\end{tabular}
}
\vspace{-4mm}

\end{table}

\vspace{-2mm}
\subsection{Visual Experience Distillation}
\label{subsec:visual_experience_distillation}

\method{} uses visual experience as training-only privileged context. 
The student receives the normal inference context \(c(x)\), while the teacher view receives the same context patched with retrieved experience:
\begin{equation}
  c_E(x)=\operatorname{Patch}(c(x),M_x).
\end{equation}
The teacher does not generate a separate trajectory; it re-scores the same student-sampled tokens under \(c_E(x)\), so inference remains unchanged while training obtains dense token-level guidance. 
Since student and teacher score the same tokens under different contexts, we draw inspiration from Skill-SD and KL-estimator analyses~\citep{skillsd,schulman2020kl,tang2025kl,liu2025kl} and use an importance-weighted sampled-token reverse-KL objective for stable cross-context distillation.

For token \(y_{i,t}\), define the student probability and the detached experience-conditioned teacher probability as
\begin{equation}
p_{i,t}=\pi^S_{\theta}(y_{i,t}\mid h_{i,t}),\qquad
q_{i,t}=\operatorname{sg}\!\left[
\pi^E_{\theta}(y_{i,t}\mid \tilde{h}_{i,t})
\right],
\end{equation}
and let \(\ell_{i,t}=\log p_{i,t}-\log q_{i,t}\). 
The sampled-token reverse-KL estimator is
\begin{equation}
  k_3(\ell_{i,t})=\exp(-\ell_{i,t})-1+\ell_{i,t}.
\end{equation}
Because tokens are sampled by the old student policy under the plain context, the SDL term uses the on-policy importance ratio
\begin{equation}
  \rho^{\mathrm{on}}_{i,t}=
  \min\!\left(
  \frac{\pi^S_{\theta}(y_{i,t}\mid h_{i,t})}
  {\operatorname{sg}[\pi^S_{\theta_{\mathrm{old}}}(y_{i,t}\mid h_{i,t})]},
  \rho_{\max}
  \right).
\end{equation}
The experience-conditioned SDL loss is
\begin{equation}
\mathcal{L}_{\mathrm{SDL}}=
\frac{1}{\sum_{i,t}m^E_{i,t}}
\sum_{i,t}m^E_{i,t}
\min\!\left(
\rho^{\mathrm{on}}_{i,t}k_3(\ell_{i,t}),c_{\mathrm{tok}}
\right),
\end{equation}
where \(m^E_{i,t}\) selects assistant tokens with non-empty retrieved experience. 
The final objective is
\begin{equation}
  \mathcal{L}_{\mathrm{GenEvolve}}=
  \mathcal{L}_{\mathrm{GRPO}}+
  \lambda_{\mathrm{SDL}}\mathcal{L}_{\mathrm{SDL}}.
\end{equation}
Unlike OPSD-style methods~\cite{opsd,opcd} that rely on privileged answers or reasoning traces, \method{} distills visual experience from best-worst generated trajectories to improve tool use, knowledge activation, reference selection, and prompt-reference synthesis.
Together with GRPO, this closes the self-evolving loop: the updated student produces stronger trajectories, yielding better visual experience for later updates.
In practice we further restrict $m^E_{i,t}$ to the agent's crucial tokens, within each sequence, keep only the top $10\%$ ranked by $|\log\pi^E_{\theta}-\log\pi^S_{\theta}|$; this concentrates the teacher signal on the few choices where the experience-conditioned policy disagrees most with the student.
Full SDL hyper-parameters and a token-level case study are provided in Appendix~\ref{subsec:sdl_details} (Figure~\ref{fig:sdl_token_evidence}, Table~\ref{tab:appendix_rl}).

\vspace{-2mm}
\section{Experiments}

We evaluate \method{} on GenEvolve-Bench using the same held-out prompts and judge rubrics. 
The main comparison covers direct image generators, recent agent frameworks, and \method{} paired open and strong downstream generators. 

\begin{figure*}[!t]
    \centering
    \includegraphics[width=14cm]{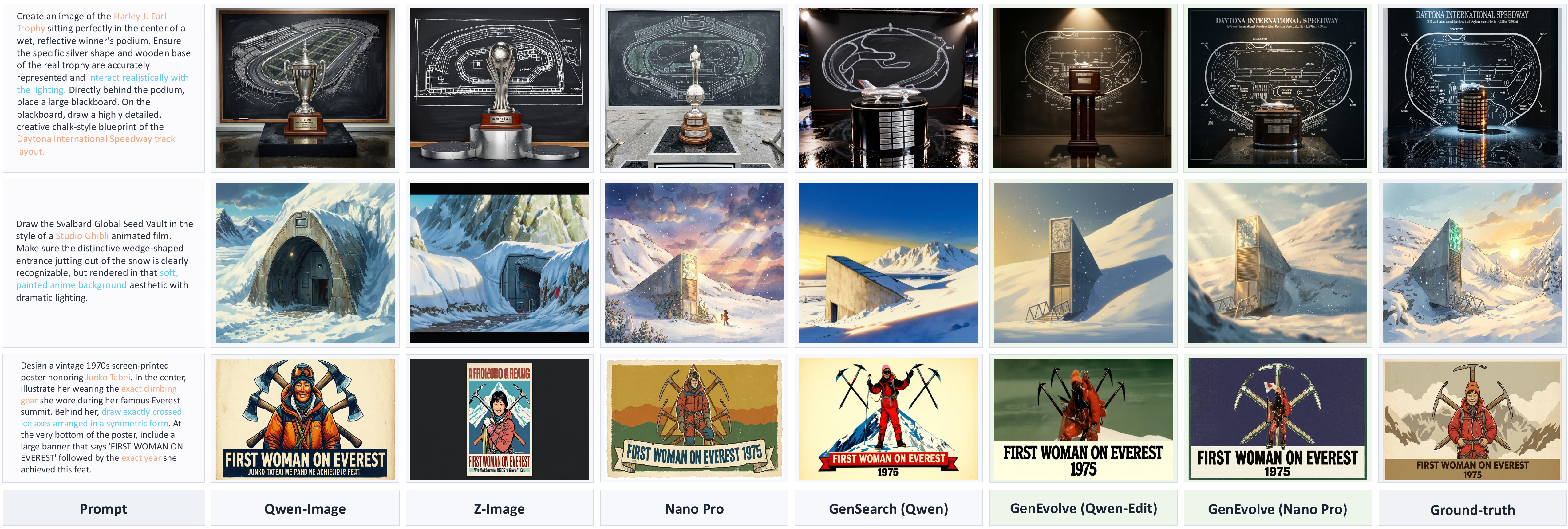}
   \caption{\textbf{Visual comparison on representative GenEvolve-Bench cases.}
Orange marks external or uncommon knowledge requirements, while blue marks internal generation-knowledge requirements; \method{} substantially improves both Qwen-based and Nano Banana Pro generation frameworks.}
    \label{fig:visual}
\vspace{-2mm}

\end{figure*}

\begin{table}[h]
\centering
\footnotesize
\caption{External evaluation on WISE~\citep{wise}. Scores are WiScore by category; higher is better. Baselines include direct generators and agentic image-generation systems. \method{} is evaluated with the same rollout-and-generate pipeline used for the open-generator setting.}
\label{tab:wise_results}
\setlength{\tabcolsep}{3.5pt}
\renewcommand{\arraystretch}{1.08}
\resizebox{\linewidth}{!}{%
\begin{tabular}{lccccccc}
\toprule
\rowcolor{GEBlue!10}
Model & Cultural & Time & Space & Biology & Physics & Chemistry & Overall \\
\midrule
\multicolumn{8}{l}{\textit{Direct generator baselines \smallpill{GEMuted}{raw}}} \\
Emu3~\cite{emu3} & 0.34 & 0.45 & 0.48 & 0.41 & 0.45 & 0.27 & 0.39 \\
\rowcolor{GELight!55}
FLUX.1-schnell~\cite{blackforestlabs2024flux1schnell} & 0.39 & 0.44 & 0.50 & 0.31 & 0.44 & 0.26 & 0.40 \\
SD-3-Medium~\cite{esser2024sd3} & 0.42 & 0.44 & 0.48 & 0.39 & 0.47 & 0.29 & 0.42 \\
\rowcolor{GELight!55}
SD-3.5-Medium~\cite{stabilityai2024sd35large} & 0.43 & 0.50 & 0.52 & 0.41 & 0.53 & 0.33 & 0.45 \\
SD-3.5-Large~\cite{stabilityai2024sd35large} & 0.44 & 0.50 & 0.58 & 0.44 & 0.52 & 0.31 & 0.46 \\
\rowcolor{GELight!55}
FLUX.1-dev~\cite{flux2024} & 0.48 & 0.58 & 0.62 & 0.42 & 0.51 & 0.35 & 0.50 \\
Hunyuan-Image 3.0~\cite{cao2025hunyuanimage3} & 0.58 & 0.57 & 0.70 & 0.56 & 0.63 & 0.31 & 0.57 \\
\rowcolor{GELight!55}
UniWorld-V2~\cite{li2025uniworldv2} & 0.60 & 0.61 & 0.70 & 0.53 & 0.64 & 0.32 & 0.58 \\
Qwen-Image~\cite{qwenimage} & 0.62 & 0.63 & 0.77 & 0.57 & 0.75 & 0.40 & 0.62 \\
\rowcolor{GELight!55}
NextFlow-RL~\cite{zhang2026nextflow} & 0.63 & 0.63 & 0.77 & 0.58 & 0.67 & 0.39 & 0.62 \\
LongCat-Image~\cite{meituan2025longcatimage} & 0.66 & 0.61 & 0.72 & 0.66 & 0.72 & 0.49 & 0.65 \\
\rowcolor{GELight!55}
DeepGen1.0~\cite{wang2026deepgen10} & 0.72 & \textbf{0.81} & 0.70 & 0.67 & 0.82 & 0.66 & 0.73 \\
GPT-4o~\cite{openai2025gpt4oimagegeneration} & 0.81 & 0.71 & \textbf{0.89} & \textbf{0.83} & 0.79 & 0.74 & 0.80 \\
\midrule
\multicolumn{8}{l}{\textit{Agentic image-generation workflows \smallpill{GEBlue}{agent}}} \\
GenAgent~\cite{jiang2026genagent} & 0.78 & 0.67 & 0.78 & 0.72 & 0.77 & 0.55 & 0.72 \\
\rowcolor{GELight!55}
Gen-Searcher-8B + Qwen-Image~\cite{gensearcher} & 0.80 & 0.71 & 0.82 & 0.76 & 0.74 & 0.75 & 0.77 \\
Mind-Brush~\cite{he2026mind} & 0.83 & 0.69 & 0.84 & 0.71 & \textbf{0.85} & 0.68 & 0.78 \\
\rowcolor{GEGreen!10}
\method{} + Qwen-Image-Edit & \textbf{0.84} & 0.74 & 0.87 & \textbf{0.83} & 0.81 & \textbf{0.83} & \textbf{0.82} \\
\bottomrule
\end{tabular}}
\end{table}

\noindent\textbf{Setup.}
The agent backbone is Qwen3-VL-8B-Instruct~\citep{qwen3vl}. 
We first cold-start it with SFT on the supervised split of GenEvolve-Data using LLaMA-Factory~\citep{llamafactory}, and then perform on-policy self-evolution on the self-evolution split (full SFT and GRPO+SDL hyper-parameters are listed in Appendix~Tables~\ref{tab:appendix_sft} and~\ref{tab:appendix_rl}).
For each prompt, the agent samples six trajectories, each producing a prompt-reference program \(z=(g,R)\).
We use Qwen-Image-Edit as the default open-source generator and Nano Banana Pro as a representative strong proprietary generator, allowing us to evaluate both open-generator performance and transfer to stronger closed-source generation.
Generated images are scored by a KScore-style visual judge, and the final programs are scored by a program-sufficiency text judge.
The mixed reward drives GRPO, while best-worst trajectory pairs provide visual experience for distillation.

All methods are evaluated on the held-out GenEvolve-Bench prompts with the same judge rubric.
We report KScore and its four dimensions: faithfulness, visual correctness, text accuracy, and aesthetics.
Full rubric weights, judge backbone, and calibration settings are listed in Appendix~\ref{subsec:reward_rubric} and Table~\ref{tab:eval_rubric}.
To assess transfer beyond our constructed benchmark, we additionally evaluate \method{} on the publicly released WISE benchmark~\citep{wise}, a commonly used knowledge-intensive image-generation benchmark; results are presented in Section~\ref{sec:wise_results} and protocol details are provided in Appendix~\ref{subsec:wise_eval}.

\paragraph{Main comparisons.}
Table~\ref{tab:main_results} reports the main comparison on GenEvolve-Bench. 
Since the benchmark includes both Knowledge-Anchored and Quality-Anchored prompts, strong performance requires factual grounding, reference use, and quality-sensitive prompt-reference construction, not only generic image quality. 
Direct generators remain limited: Qwen-Image reaches only 0.2987 KScore, while the stronger Nano Banana Pro reaches 0.5298 but still benefits from agent-side orchestration on demanding grounded cases.
Under the same open-generator setting, \method{} improves over Gen-Searcher, a recent trained search-grounded agentic baseline, raising KScore from 0.3493 to 0.3663 and visual correctness from 0.1050 to 0.1338.
Although Qwen-Image-Edit is not a particularly strong renderer, this gain shows the benefit of better agent-side orchestration, including external search, internal knowledge activation, reference selection, and prompt-reference synthesis. 
With the stronger Nano Banana Pro, the learned agent is further unlocked: \method{} reaches the best overall KScore of 0.5739 and improves most dimensions over both raw Nano Banana Pro and Gen-Searcher with the same generator.
This suggests generator-transferable orchestration rather than overfitting to one renderer. 
Figure~\ref{fig:visual} provides representative visual comparisons.

\paragraph{External Generalization on WISE.}
\label{sec:wise_results}
To verify that the policy learned by \method{} generalizes beyond our constructed benchmark, we evaluate it on the publicly released WISE benchmark~\citep{wise}, which targets knowledge-intensive image generation across six categories: cultural, time, space, biology, physics, and chemistry. Following the standard protocol, the agent receives only the WISE prompt, produces a prompt-reference program through the normal rollout interface, and the output image is generated by Qwen-Image-Edit. Generated images are scored under the WISE three-dimension protocol with GPT-4o-2024-05-13 as the judge, and results are aggregated by the official WiScore script.

Table~\ref{tab:wise_results} compares \method{} with both direct generators and recent agentic image-generation systems. \method{} attains the best overall WiScore of \textbf{0.82}, surpassing the strongest direct baseline (GPT-4o, 0.80) and all agentic baselines (e.g., GenAgent 0.72, Gen-Searcher-8B 0.77, Mind-Brush 0.78). The improvement is most pronounced on \emph{chemistry} (0.83) and \emph{biology} (0.83), where factual grounding through tool-orchestrated trajectories is most beneficial. Figure~\ref{fig:teaser}\,(b) further visualizes these category-wise gains. These results show that the tool-orchestrated visual trajectory and Visual Experience Distillation transfer to an out-of-distribution knowledge-intensive benchmark, rather than overfitting to GenEvolve-Bench prompts or judges.

\paragraph{Ablation study.}
Table~\ref{tab:main_ablation} studies the contribution of each training stage. 
Direct Qwen-Image reaches 0.2987 KScore, and an untuned Qwen3-VL workflow improves to 0.3317 by using the same tool interface and reference-conditioned generation.
SFT cold start further raises KScore to 0.3480, showing that curated trajectories teach more reliable tool use and prompt-reference construction.
Adding GRPO without visual experience improves KScore to 0.3548, but scalar rewards alone provide limited credit assignment for long tool trajectories.
The full \method{} reaches 0.3663, with the highest visual correctness of 0.1338 and the best scores on both Knowledge-Anchored and Quality-Anchored tracks.
This shows that Visual Experience Distillation provides complementary token-level guidance beyond SFT and scalar-reward GRPO.
% \noindent\textbf{Observed trends.}
% The results expose the behavior that \method{} is designed to improve. First, direct generators struggle on tool-demanding requests even when their general image quality is high: Nano Banana Pro is the strongest direct generator with a KScore of 0.5298, while open direct generators remain below 0.30. Their main weakness is visual correctness under grounding and verifiable quality constraints; Qwen-Image reaches only 0.0623 on this dimension. Second, agentic generation helps when the final generator is fixed. With Qwen-Image-Edit as the downstream generator, \method{} reaches 0.3663 KScore and 0.1338 visual correctness, outperforming the generic tool workflow and the search-augmented baseline on overall KScore. Third, the improvement is not isolated to one track: \method{} improves both Knowledge-Anchored and Quality-Anchored prompts. Finally, the downstream generator remains a major bottleneck. When the same \method{} agent output is generated by Nano Banana Pro rather than Qwen-Image-Edit, KScore rises to 0.5739, indicating that the agent policy can construct useful generation programs whose final quality is partly limited by reference-conditioned generation fidelity.

\begin{table}[t]
\centering
\footnotesize

\caption{Component ablation on GenEvolve-Bench. Except for the generator-only row, all variants use Qwen-Image-Edit as the downstream generator.}
\label{tab:main_ablation}
\setlength{\tabcolsep}{3.5pt}
\renewcommand{\arraystretch}{1.10}
\resizebox{\linewidth}{!}{%
\begin{tabular}{p{0.34\linewidth}ccccccc}
\toprule
\rowcolor{GEAmber!10}
Variant & KScore (All) & Faith. & Vis. & Text & Aesth. & Know.-Anch. & Qual.-Anch. \\
\midrule
\smallpill{GEMuted}{raw} Qwen-Image only & 0.2987 & 0.3729 & 0.0623 & 0.4226 & \textbf{0.6751} & 0.2384 & 0.3768 \\
\rowcolor{GELight!55}
\smallpill{GEMuted}{base} Untuned Qwen3-VL workflow & 0.3317 & 0.4781 & 0.1061 & 0.4571 & 0.5867 & 0.3109 & 0.3587 \\
SFT only 
& 0.3480 & 0.4791 & 0.1121 & \textbf{0.4934} & 0.5785 & 0.3303 & 0.3709 \\
\rowcolor{GELight!55}
SFT + GRPO w/o visual experience
& 0.3548 & 0.5027 & 0.1138 & 0.4926 & 0.6197 & 0.3277 & 0.3898 \\
\rowcolor{GEGreen!8}
\smallpill{GEGreen}{full} \method{} & \textbf{0.3663} & \textbf{0.5303} & \textbf{0.1338} & 0.4907 & 0.6347 & \textbf{0.3410} & \textbf{0.3990} \\
\bottomrule
\end{tabular}}
\vspace{-4mm}
\end{table}

\vspace{-2mm}
\section{Conclusion}
% \vspace{-0.1cm}
We present \method{}, a self-evolving framework for image-generation agents via Tool-Orchestrated Visual Experience Distillation. Instead of a single prompt-to-image call, \method{} formulates generation as a visual trajectory where an agent coordinates external evidence, visual references, internal generation knowledge, prompt-reference construction, generation, and feedback. By comparing best-worst trajectories, \method{} extracts structured visual experience and distills this teacher-only guidance into the student policy. Experiments on GenEvolve-Bench show improved agentic behavior and final image quality, demonstrating its value compared with traditional generation.

\clearpage

\nocite{betker2023improving,deepseekmath}
\bibliographystyle{plainnat}
\bibliography{references}

\clearpage

\appendix

\section{GenEvolve-Data Construction}
\label{sec:appendix_data_construction}

\subsection{Prompt Pool Recipes}

GenEvolve-Data is constructed to support three stages of our framework: supervised cold start, self-evolution, and held-out evaluation. 
Unlike ordinary prompt-rewriting data, each example is designed as a complete visual generation problem in which an agent must acquire missing evidence, select visual references, activate generation knowledge when needed, and synthesize a prompt-reference program. 
We therefore begin from recipe-controlled prompt generation rather than unconstrained LLM sampling. 
Each recipe specifies the task family, missing factual information, expected visual anchor, dominant visible requirement, optional secondary constraints, and difficulty. 
These fields are used for coverage control and auditing, but are not exposed to the agent as task labels.

\begin{table}[h]
\centering
\footnotesize
\caption{Prompt recipe fields used to construct the prompt pool.}
\label{tab:appendix_recipes}
\setlength{\tabcolsep}{4pt}
\renewcommand{\arraystretch}{1.12}
\resizebox{\linewidth}{!}{%
\begin{tabular}{p{0.18\linewidth} p{0.30\linewidth} p{0.44\linewidth}}
\toprule
\rowcolor{GEBlue!10}
Field & Role & Examples \\
\midrule
Task family 
& Controls the high-level generation scenario 
& factual scene, text-critical generation, multi-object layout, body/anatomy, attribute binding, material, creative transfer, mixed hard case \\

\rowcolor{GELight!55}
Factual gap 
& Specifies information that may require search 
& entity names, official labels, dates, variants, visual text, event-specific details \\

Visual anchor 
& Specifies appearance cues that benefit from image search 
& entity shape, facade, costume, product geometry, pose, material, style \\

\rowcolor{GELight!55}
Dominant requirement 
& Defines the main visible quality challenge 
& text rendering, layout, counting, material, anatomy, aesthetics \\

Secondary requirements 
& Adds realistic multi-constraint composition 
& additional object relations, style constraints, reference binding, readable text \\

\rowcolor{GELight!55}
Difficulty 
& Controls expected trajectory depth 
& medium or hard; rare easy prompts are retained only after validation \\
\bottomrule
\end{tabular}}
\end{table}

The current pool contains 19,990 valid prompts after deduplication: 11,999 Knowledge-Anchored prompts and 7,991 Quality-Anchored prompts. 
The average prompt length is about 65 words, with 13,333 hard prompts, 6,654 medium prompts, and 3 easy prompts. 
Table~\ref{tab:appendix_data_config} summarizes the concrete configuration used by the current implementation.

\begin{table}[h]
\centering
\footnotesize
\caption{Concrete GenEvolve-Data configuration.}
\label{tab:appendix_data_config}
\setlength{\tabcolsep}{4pt}
\renewcommand{\arraystretch}{1.12}
\resizebox{\linewidth}{!}{%
\begin{tabular}{p{0.22\linewidth} p{0.24\linewidth} p{0.46\linewidth}}
\toprule
\rowcolor{GEBlue!10}
Component & Value & Notes \\
\midrule
Prompt pool 
& 19,990 valid prompts 
& Constructed from 19,991 pre-dedup valid candidates; 9 invalid candidates removed by length filtering. \\

\rowcolor{GELight!55}
Prompt tracks 
& 11,999 / 7,991 
& Knowledge-Anchored / Quality-Anchored prompts. \\

Categories 
& 16 categories 
& Architecture, city streets, public figures, products, vehicles, events, science, artifacts, text/layout, spatial, counting, anatomy, attributes, material, aesthetics, creative transfer. \\

\rowcolor{GELight!55}
Search labels 
& 19,990 image search; 13,972 text search 
& All prompts require image search. Text search includes all Knowledge-Anchored prompts and 1,973 Quality-Anchored prompts. \\

Dominant quality tags 
& 8 visible requirement types 
& Material 3,000; creative transfer 3,000; text 3,000; anatomy 3,000; attribute binding 2,999; counting 2,995; spatial layout 1,000; aesthetics 996. \\

\rowcolor{GELight!55}
Difficulty 
& 13,333 hard; 6,654 medium; 3 easy 
& Difficulty controls expected search/reference/knowledge-activation depth. \\

Self-evolution/eval format 
& \texttt{\{id, question, gt\_image\}} 
& Teacher trajectory, selected references, and teacher prompt are removed from the rollout input. \\

\rowcolor{GELight!55}
Evaluation metadata 
& track, category, difficulty, search flags 
& Also includes expected reference targets and quality-requirement tags for analysis. \\
\bottomrule
\end{tabular}}
\end{table}

\subsection{Teacher Trajectory Generation}

Each validated prompt is converted into a teacher trajectory through a real multi-turn tool loop. 
At each turn, the teacher emits reasoning and either one tool call or a final answer. 
The system executes the tool, appends the observation, and continues until a valid prompt-reference program is produced or the trajectory times out. 
We use strong multimodal teacher models, including Seed2.0 and Gemini 3 Pro, to generate these trajectories because they provide strong reasoning, reference understanding, and tool-use capabilities~\citep{seed2026seed2,google2025gemini3pro}.

Accepted trajectories must contain meaningful tool use rather than only a final prompt. 
Image search is required because the downstream generator is reference-conditioned. 
Generation-knowledge calls are encouraged when the prompt contains visible quality challenges, but the system does not force every trajectory to query all knowledge types. 
This preserves realistic tool-order diversity: some trajectories begin with factual lookup, some start from reference search, and others activate generation knowledge only after inspecting the request and retrieved evidence.

\subsection{Filtering Rubric}

Filtering combines hard programmatic checks and VLM-based judgment. 
Hard checks reject incomplete trajectories, missing image search, invalid reference counts, invalid local image paths, unparseable JSON, invalid skill names, missing ordinal reference wording, meaningless reasoning, unsafe content, and raw URL leakage in the final prompt. 
These rules remove format and protocol failures before semantic review.

The VLM filter then scores six dimensions: prompt suitability, reference grounding, trajectory process quality, skill integration, final prompt faithfulness, and supervised training value. 
A trajectory is kept only if the final program remains faithful to the user request, the selected references support the claimed visual details, and the collected evidence is actually used. 
In the full run, 13,379 of 19,320 structurally valid trajectories were kept (69.2\%). 
Average skill integration was 4.70/5, while reference grounding was the hardest dimension at 3.98/5. 
Common failures include hallucinated reference content, contradicted image details, duplicate references, unused tool results, and final prompts that copy the user request without meaningful synthesis.

\subsection{GT Image Generation and Filtering}

For self-evolution and evaluation, we render GT images from high-quality teacher prompt-reference programs and their selected references. 
We use Nano Banana Pro as the GT image generator because of its strong instruction following, reference-conditioned editing, text rendering, and real-world visual knowledge~\citep{google2025nanobananapro}. 
These GT images are not unique ground-truth answers to the raw user prompts; instead, they provide strong visual realizations that make image-level feedback and evaluation meaningful.

We generated 4,321 successful GT images from 4,379 attempts (98.7\%) and retained 3,175 after filtering (73.5\%). 
The image filter checks generation-prompt compliance, reference utilization, visual coherence, and image quality. 
Images that ignore selected references, fail required text, contradict grounded details, or exhibit severe visual artifacts are discarded.

\subsection{Supervised and Self-Evolution Export}

GenEvolve-Data is exported into two complementary views. 
The supervised view preserves the full tool-loop conversation and all images shown to the model, including candidate references, so that the student learns evidence acquisition, candidate comparison, reference selection, internal-knowledge activation, and prompt-reference program construction. 
During supervised training, only assistant-side tokens are optimized; user prompts and tool observations serve as context and are masked from the loss. 
The supervised split contains 8,800 training examples and 200 evaluation examples.

The self-evolution view removes the teacher trajectory and teacher final program. 
Each example contains the raw prompt, GT image path, and metadata needed for reward evaluation. 
This prevents the agent from copying teacher actions and forces it to produce its own tool-orchestrated rollout. 
The current split uses 3,175 filtered GT image cases: a 2,575-case self-evolution training pool and a about 600-case evaluation pool. 
The training pool is further divided into 2,446 optimization cases and 129 internal validation cases.

\subsection{Coverage and Construction Statistics}

\begin{figure}[h]
\centering
\includegraphics[width=0.58\linewidth]{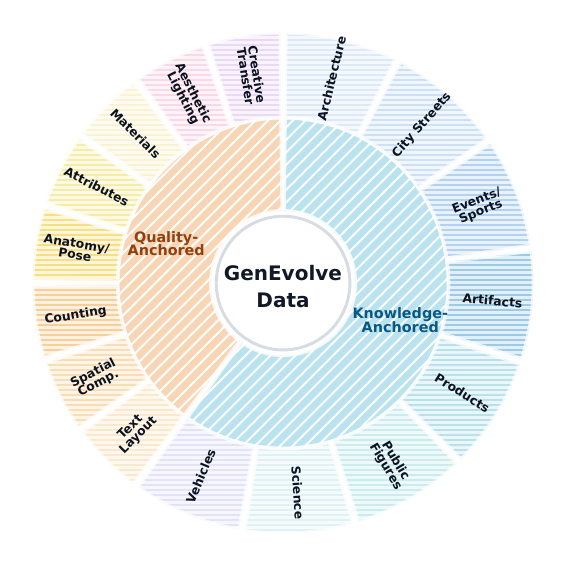}
\caption{\textbf{Category hierarchy of GenEvolve-Data.} The prompt pool is organized into Knowledge-Anchored and Quality-Anchored tracks, each covering eight diagnostic categories used for coverage control, split stratification, and benchmark analysis.}
\label{fig:data_sunburst}
\end{figure}

\begin{figure}[h]
\centering
\includegraphics[width=\linewidth]{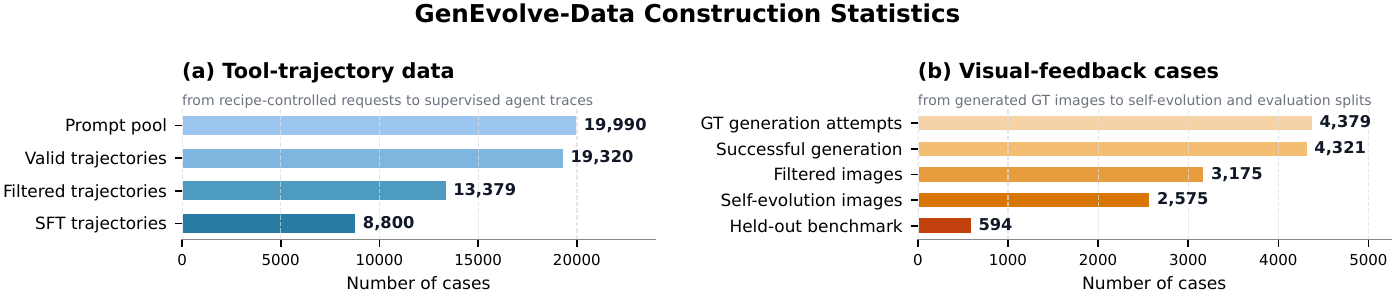}
\caption{\textbf{GenEvolve-Data construction statistics.} The left panel summarizes prompt-to-trajectory filtering for supervised learning, and the right panel summarizes GT image generation, image filtering, self-evolution images, and held-out benchmark cases.}
\label{fig:data_statistics}
\end{figure}

Figure~\ref{fig:data_sunburst} visualizes the two-track category hierarchy, and Figure~\ref{fig:data_statistics} summarizes the major filtering and split statistics. 
Together, these figures show that GenEvolve-Data covers both externally grounded generation and quality-sensitive generation requirements, while maintaining a held-out benchmark split with no exact overlap with the self-evolution training pool.

\section{Additional Method Details}

This section provides implementation details for the rollout protocol, prompt-reference program schema, experience memory, retrieval, GRPO loss, and experience-conditioned self-distillation. These details complement the main method while keeping the core paper concise.

\subsection{Tool-Orchestrated Rollout Protocol}
\label{subsec:tool_protocol}

At inference time, \method{} exposes a small and auditable action space. Each assistant turn must either emit one tool call or terminate with a parseable final answer. The environment executes the requested tool, appends the observation, and resumes the agent. The current runtime uses the tools in Table~\ref{tab:appendix_tools}.

\begin{table}[h]
\centering
\footnotesize
\caption{Runtime tools used by the image-generation agent.}
\label{tab:appendix_tools}
\setlength{\tabcolsep}{4pt}
\renewcommand{\arraystretch}{1.12}
\resizebox{\linewidth}{!}{%
\begin{tabular}{p{0.18\linewidth} p{0.23\linewidth} p{0.31\linewidth} p{0.20\linewidth}}
\toprule
\rowcolor{GEBlue!10}
Tool & Input & Observation & Main role \\
\midrule
\tool{search} 
& Textual query 
& Snippets for names, dates, labels, event facts, official terms, and other visible facts 
& External factual evidence \\

\rowcolor{GELight!55}
\tool{image\_search} 
& Visual query 
& Candidate images with IDs, titles, source URLs, cached paths, and previews 
& Visual reference retrieval \\

\tool{query\_knowledge} 
& One enum skill name 
& Static generation guidance for the requested knowledge type 
& Internal generation-knowledge activation \\
\bottomrule
\end{tabular}}
\end{table}

The agent is not forced to call tools in a fixed order. Knowledge-Anchored prompts often use \tool{search} before \tool{image\_search}; Quality-Anchored prompts may skip text search when the visual anchor is sufficient; and complex prompts may call several generation-knowledge tools. The final answer must select one or two reference images and synthesize a prompt-reference program. Invalid tool names, invalid knowledge names, missing references, raw URL leakage, and unparseable final JSON are treated as trajectory failures during data construction and evaluation.

\begin{table}[h]
\centering
\footnotesize
\caption{Callable internal generation-knowledge modules. We instantiate them using the common ``skill'' interface, but conceptually they serve as on-demand generation knowledge for visible failure modes.}
\label{tab:appendix_skills}
\setlength{\tabcolsep}{3pt}
\renewcommand{\arraystretch}{1.12}
\resizebox{\linewidth}{!}{%
\begin{tabular}{p{0.20\linewidth} p{0.27\linewidth} p{0.20\linewidth} p{0.27\linewidth}}
\toprule
\rowcolor{GEAmber!10}
Module & Visible failure mode & Module & Visible failure mode \\
\midrule
\smallpill{GEBlue}{text} \skill{text rendering} 
& unreadable, misplaced, or misspelled text 
& \smallpill{GECyan}{layout} \skill{spatial layout} 
& wrong positions, depth, scale, or occlusion \\

\rowcolor{GELight!55}
\smallpill{GERed}{body} \skill{anatomy/body coherence} 
& malformed hands, pose, face, or body structure 
& \smallpill{GEAmber}{style} \skill{aesthetic drawing} 
& weak lighting, camera, composition, or atmosphere \\

\smallpill{GEGreen}{bind} \skill{attribute binding} 
& attributes leak between objects 
& \smallpill{GEBlue}{create} \skill{creative drawing} 
& weak style transfer or concept fusion \\

\rowcolor{GELight!55}
\smallpill{GEAmber}{physics} \skill{physical/material consistency} 
& impossible support, shadow, reflection, or material behavior 
& \smallpill{GERed}{count} \skill{quantity counting} 
& wrong number of instances or uncountable layout \\
\bottomrule
\end{tabular}}
\end{table}

\subsection{Prompt-Reference Program Schema}

The final executable object is a prompt-reference generation program \(z=(g,R)\). The instruction \(g\) is natural language consumed by a reference-conditioned generator, while \(R\) is an ordered list of local reference images. The instruction must refer to selected references by ordinal phrases such as ``the first reference image'', rather than raw image IDs or URLs. This makes the program independent of transient retrieval IDs and aligns it with generator APIs that receive images as ordered inputs.

\begin{verbatim}
{
  "gen_prompt": "... the first reference image ...",
  "reference_images": [
    {"img_id": "IMG_001", "role": "identity/shape reference"},
    {"img_id": "IMG_004", "role": "style or layout reference"}
  ],
  "selected_skills": ["spatial_layout", "text_rendering"],
  "rationale": "Short explanation of evidence, references, and knowledge."
}
\end{verbatim}

The selected skills are not passed to the generator as separate hidden controls. They record which internal generation knowledge was queried and should be reflected in \(g\). During training, this field supports tool-call supervision, diagnostic analysis, and visual-experience slot construction.

\subsection{Visual Experience Extraction --- Real Training Cases}
\label{subsec:real_experience_case}

To make the experience extraction process concrete, we present three complete best-worst trajectory comparisons extracted during self-evolution training. For each case, we show the original user request, the full tool-call sequence and thinking excerpts from both the best and worst trajectories, the generated images, and the five experience slots distilled from the comparison. All content is taken verbatim from the training logs; only tool observations are abbreviated for space.

These cases illustrate three distinct failure modes that Visual Experience Distillation is designed to address. \textbf{Case 1} demonstrates how a subtle search query mistake --- including an attribute keyword that introduces noise --- propagates into a completely wrong factual grounding and ruins the entire generation. \textbf{Case 2} shows that retrieving correct facts is necessary but insufficient: skipping generation-knowledge skills causes the agent to produce a syntactically valid but visually broken prompt, because text layout and typographic structure require explicit skill guidance that cannot be replaced by a style reference image. \textbf{Case 3} shows how omitting one critical skill (\texttt{spatial\_layout}) while correctly calling others leads to cascading layout and attribute failures --- when two distinct objects must be spatially separated, vague positioning descriptions cause the generator to merge or overlap them.

Together, these cases demonstrate that the experience extraction mechanism captures actionable lessons across the full trajectory decision space: search query formulation (Case~1), skill routing and activation (Cases~2--3), reference selection (all cases), prompt-reference program synthesis (all cases), and failure avoidance patterns (all cases). The extracted experience slots are then injected only into the privileged teacher branch during SDL, providing dense token-level guidance that complements the scalar reward signal from GRPO.

\vspace{2mm}
\begin{tcolorbox}[
  colback=GELight,
  colframe=GEMuted,
  coltitle=white,
  fonttitle=\bfseries\small,
  title=User Request,
  sharp corners,
  boxrule=0.5mm,
  left=2mm, right=2mm, top=1.5mm, bottom=1.5mm
]
\small
\textit{``Draw a futuristic, cyberpunk version of the Crucible Theatre's main stage setup used for the World Snooker Championship. The iconic, intimate circular seating layout should remain recognizable. The snooker table should be centered, constructed from polished dark mahogany and thick, glowing transparent glass. The table felt must match the national flag stripes of the 2023 World Snooker Champion. Above the table, add a translucent holographic scoreboard displaying `147' in bright cyan digits.''}
\end{tcolorbox}

\vspace{2mm}
\begin{tcolorbox}[
  colback=GEGreenLight,
  colframe=GEGreen,
  coltitle=white,
  fonttitle=\bfseries\small,
  title=Best Trajectory \quad R=1.00 \quad (image: 1.0 / text: 1.0),
  sharp corners,
  boxrule=0.5mm,
  left=2mm, right=2mm, top=1.5mm, bottom=1.5mm,
  breakable
]
\footnotesize
\textbf{Turn 1 --- Think \& Tool Call:} \texttt{search("2023 World Snooker Championship winner nationality")}\\
\textit{``\ldots First, I need to find out who won the 2023 World Snooker Championship, right? Because the table felt has to match their national flag's stripes. So first, use the search tool for that query\ldots''}

\vspace{1mm}
\textbf{$\rightarrow$ Observation:} Search returns \textbf{Luca Brecel} from \textbf{Belgium}.

\vspace{1mm}
\textbf{Turn 2 --- Think \& Tool Call:} \texttt{image\_search("Crucible Theatre Sheffield World Snooker Championship main stage setup circular seating")}\\
\textit{``\ldots the 2023 winner is Luca Brecel from Belgium, so his national flag is the Belgian flag, which has vertical black, yellow, red stripes. Got that part. Now I need a visual reference of the Crucible Theatre main stage setup\ldots''}

\vspace{1mm}
\textbf{$\rightarrow$ Observation:} Returns images of the iconic circular seating layout. Selects \textbf{IMG\_001}.

\vspace{1mm}
\textbf{Turn 3 --- Think \& Tool Call:} \texttt{query\_knowledge("creative\_drawing")}\\
\textit{``\ldots IMG\_001 looks like it shows the actual stage setup with the circular seating, that's perfect for reference. Now, what skills do I need? 1.~Creative drawing: because we need a cyberpunk futuristic version of the real setup, keeping the circular seating recognizable\ldots''}

\vspace{1mm}
\textbf{Turn 4 --- Think \& Tool Call:} \texttt{query\_knowledge("spatial\_layout")}\\
\textit{``\ldots I need to make sure the elements are positioned correctly: circular seating in the background, snooker table centered in the midground, holographic scoreboard above the table\ldots''}

\vspace{1mm}
\textbf{Turn 5 --- Think \& Tool Call:} \texttt{query\_knowledge("text\_rendering")}\\
\textit{``\ldots the holographic scoreboard has to show `147' in bright cyan digits, so I need to make sure the text is rendered correctly\ldots''}

\vspace{1mm}
\textbf{Turn 6 --- Think \& Final Answer:}\\
\textit{``\ldots reference the first reference image for the iconic circular seating layout. Then, cyberpunk futuristic style, midground has the centered snooker table: polished dark mahogany frame, thick glowing transparent glass sides, felt matching the \textbf{vertical black-yellow-red stripes of the Belgian national flag}\ldots Above the table, centered in the upper third of the frame, a translucent holographic scoreboard displaying `147' in bold cyan digits\ldots''}

\vspace{1mm}
\textbf{Final program (gen\_prompt):} \textit{``Futuristic cyberpunk version of the Crucible Theatre main stage setup, using the iconic circular seating layout from the first reference image for recognizability. Depth layering: background = the circular seating from the first reference, reimagined with neon accent lighting in cyberpunk style; midground = centered snooker table, with polished dark mahogany frame, thick glowing transparent glass sides, and table felt matching the \textbf{vertical black-yellow-red stripes of the Belgian national flag}; foreground = subtle neon haze. Above the table\ldots a translucent holographic scoreboard displays the text \textbf{`147'} in bright bold sans-serif cyan glowing digits\ldots''}\\
\textbf{Reference:} IMG\_001 --- note: \textit{``Iconic circular seating layout of the Crucible Theatre main stage, to retain recognizability in the cyberpunk reimagining.''}
\end{tcolorbox}

\vspace{2mm}
\begin{tcolorbox}[
  colback=GERedLight,
  colframe=GERed,
  coltitle=white,
  fonttitle=\bfseries\small,
  title=Worst Trajectory \quad R=0.225 \quad (image: 0.45 / text: 0.0),
  sharp corners,
  boxrule=0.5mm,
  left=2mm, right=2mm, top=1.5mm, bottom=1.5mm,
  breakable
]
\footnotesize
\textbf{Turn 1 --- Think \& Tool Call:} \texttt{search("2023 World Snooker Championship winner national flag")}\\
\textit{``\ldots first, let's confirm the 2023 winner. Let's call search with query `2023 World Snooker Championship winner national flag'. Then, after that, do image search for Crucible Theatre main stage\ldots''}

\vspace{1mm}
\textbf{$\rightarrow$ Observation:} Search returns mixed results due to the keyword ``flag'' causing noise. The agent \textbf{incorrectly concludes} the winner is \textbf{Wu Yize from China}.

\vspace{1mm}
\textbf{Turn 2 --- Think \& Tool Call:} \texttt{image\_search("Crucible Theatre Sheffield World Snooker Championship main stage circular seating layout")}\\
\textit{``\ldots 2023 winner is Wu Yize, Chinese, so Chinese national flag has red background, five yellow stars. Got that. Now I need to get a visual reference of the Crucible Theatre main stage setup\ldots''}

\vspace{1mm}
\textbf{Turn 3 --- Think \& Tool Call:} \texttt{query\_knowledge("creative\_drawing")}\\
\textit{``\ldots IMG\_002 looks like a good visual of the actual venue seating from a fan's perspective. Now I need to query the relevant skills. First, creative\_drawing because we need a cyberpunk futuristic version of the real venue\ldots''}

\vspace{1mm}
\textbf{Turn 4 --- Think \& Tool Call:} \texttt{query\_knowledge("attribute\_binding")}\\
\textit{``\ldots we have multiple elements with specific distinct attributes: the snooker table has polished dark mahogany construction, thick glowing transparent glass edges, felt matching Chinese flag stripes (red with yellow stars)\ldots attribute binding will help make sure each attribute is correctly assigned\ldots''}

\vspace{1mm}
\textbf{Turn 5 --- Think \& Tool Call:} \texttt{query\_knowledge("text\_rendering")}\\
\textit{``\ldots there's a holographic scoreboard displaying `147' in bright cyan digits, so I need to make sure that text is rendered correctly\ldots''}

\vspace{1mm}
\textbf{Turn 6 --- Think \& Final Answer:}\\
\textit{``\ldots reference the first reference image for the Crucible Theatre intimate circular seating layout\ldots cyberpunk futuristic style: neon accents, dark moody lighting\ldots snooker table constructed from polished dark mahogany with thick glowing transparent glass edges, its felt exactly matching the \textbf{red background and five yellow star stripes of the Chinese national flag}\ldots''}

\vspace{1mm}
\textbf{Final program (gen\_prompt):} \textit{``Futuristic cyberpunk reinterpretation of the Crucible Theatre main stage\ldots Centered on the stage: a snooker table constructed from polished dark mahogany with thick glowing transparent glass edges, its felt exactly matching the \textbf{red background and five yellow star stripes of the Chinese national flag}. Centered directly above the snooker table: a translucent holographic scoreboard displaying the text `147' in bold bright cyan glowing sans-serif digits\ldots''}\\[1mm]
\textcolor{GERed}{\textbf{Critical failures:}}
(1)~The search query included the word ``flag'', causing keyword-matching noise that led to \textbf{an incorrect entity identification} (Wu Yize instead of Luca Brecel).
(2)~The entire downstream trajectory --- reference selection, skill integration, prompt construction --- was built on a \textbf{wrong factual premise}, producing Belgian $\to$ Chinese flag stripes.
(3)~Despite correct tool use for creative\_drawing, attribute\_binding, and text\_rendering, the \textbf{text reward is 0.0} because the flag constraint is factually wrong.
\end{tcolorbox}

\vspace{2mm}
\begin{tcolorbox}[
  colback=GEAmberLight,
  colframe=GEAmber,
  coltitle=white,
  fonttitle=\bfseries\small,
  title=Extracted Experience Slots \quad (Delta = 0.775 above threshold),
  sharp corners,
  boxrule=0.5mm,
  left=2mm, right=2mm, top=1.5mm, bottom=1.5mm
]
\footnotesize
\begin{itemize}[leftmargin=1.2em, itemsep=2pt, topsep=2pt]
\item[\textcolor{GEBlue}{\textbf{S1}}] \textbf{Search strategy:} When a constraint requires an attribute of a specific entity (e.g., a champion's flag), search for the entity first (``2023 World Snooker Championship winner''), then deduce the attribute --- rather than including the attribute keyword (``flag'') in the search query, which causes keyword-matching noise.
\item[\textcolor{GEAmber}{\textbf{S2}}] \textbf{Knowledge activation:} Call \texttt{spatial\_layout} to explicitly define the positioning and depth layering of the environment (circular seating) versus the central subject (snooker table), preventing the environment from being omitted.
\item[\textcolor{GECyan}{\textbf{S3}}] \textbf{Reference selection:} Explicitly map the reference image to a specific depth layer (e.g., ``background = circular seating from the first reference'') so the layout is preserved in the cyberpunk reimagining.
\item[\textcolor{GEGreen}{\textbf{S4}}] \textbf{Prompt construction:} Structure the prompt using explicit depth layering (background, midground, foreground) to separate the environment from the central subject, preventing the model from focusing entirely on the subject and omitting the venue.
\item[\textcolor{GERed}{\textbf{S5}}] \textbf{Failure avoidance:} Do not include visual attribute keywords (e.g., ``flag'') in the initial factual search query --- this biases the search engine toward irrelevant articles and can cause incorrect entity resolution.
\end{itemize}
\vspace{1mm}
% {\scriptsize\textcolor{GEMuted}{Case 1: Real training data. The search query difference (``winner nationality'' vs.\ ``winner national flag'') caused the entire trajectory to diverge.}}
\end{tcolorbox}

\begin{figure}[h]
\centering
\begin{minipage}{0.48\linewidth}
\centering
\includegraphics[width=\linewidth]{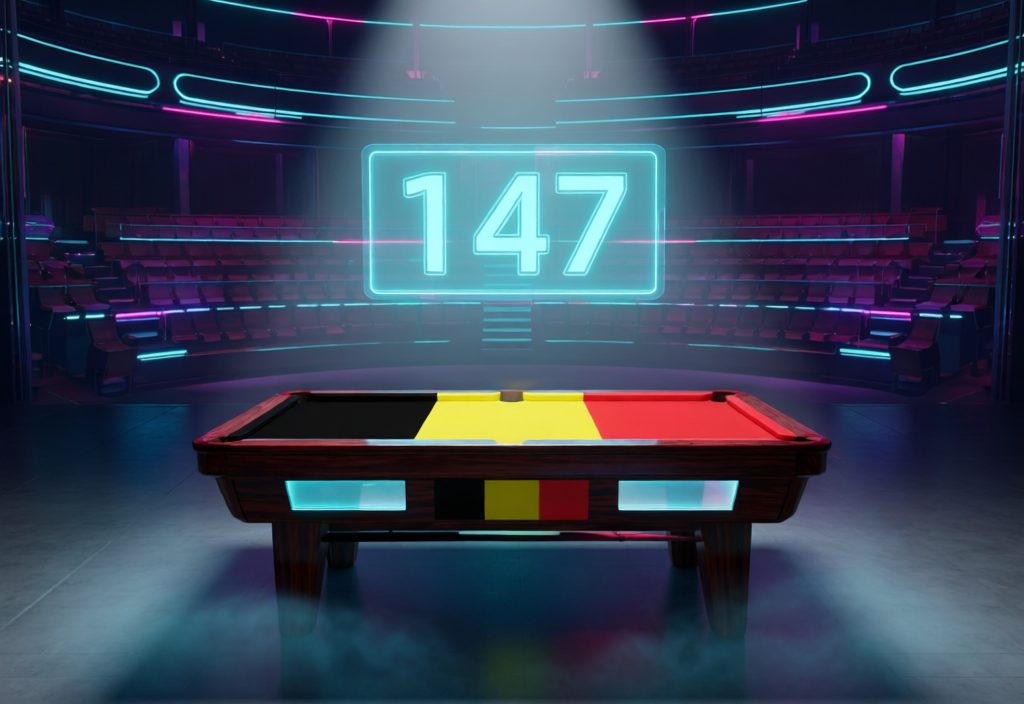}\\[2pt]
{\small\textcolor{GEGreen}{\textbf{Best} ($R=1.0$): Belgian flag stripes (correct)}}
\end{minipage}
\hfill
\begin{minipage}{0.48\linewidth}
\centering
\includegraphics[width=\linewidth]{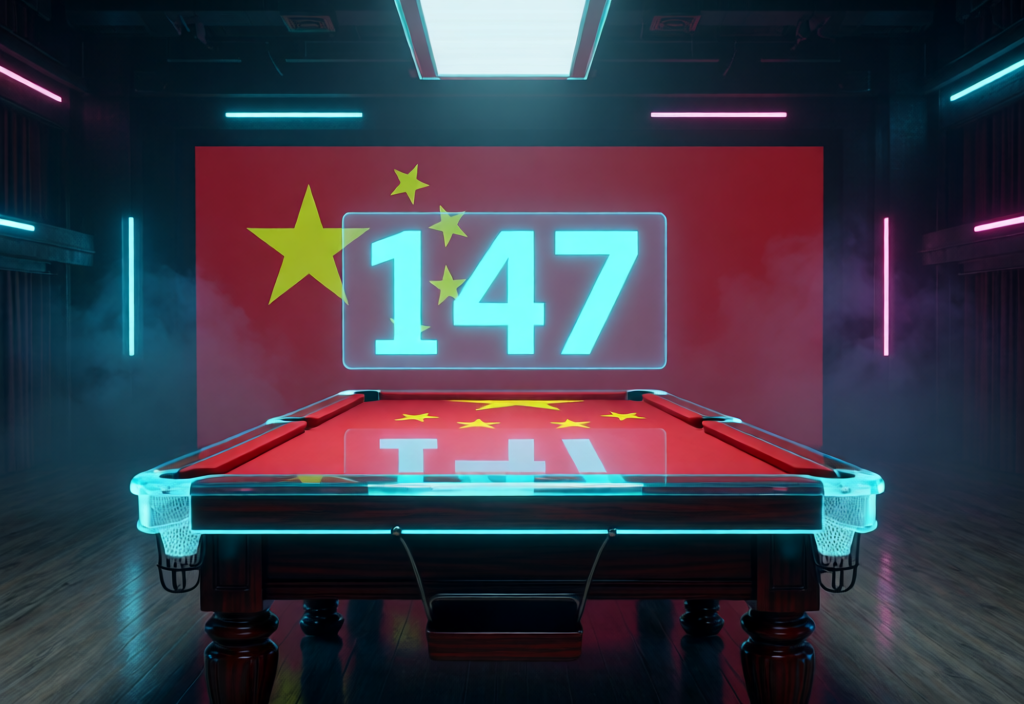}\\[2pt]
{\small\textcolor{GERed}{\textbf{Worst} ($R=0.225$): Chinese flag colors (wrong )}}
\end{minipage}
\caption{\textbf{Case 1 generated images.} The search query ``winner nationality'' (best) vs.\ ``winner national flag'' (worst) led to completely different factual grounding and flag stripe colors on the snooker table felt.}
\label{fig:case1_images}
\end{figure}

%% ═══════════════════════ CASE 2: Aérotrain ═══════════════════════

\vspace{4mm}
\begin{tcolorbox}[
  colback=GELight, colframe=GEMuted, coltitle=white, fonttitle=\bfseries\small,
  title=Case 2 --- User Request, sharp corners, boxrule=0.5mm,
  left=2mm, right=2mm, top=1.5mm, bottom=1.5mm
]
\small
\textit{``Create a retro-futuristic 1970s-style travel poster featuring the French A\'{e}rotrain I80. The poster should show the hovertrain gliding on its inverted T-shaped concrete track. In bold vintage typography, include the exact top speed it reached during its world record run, along with the year that record was set. The text should be styled like a classic transport advertisement.''}
\end{tcolorbox}

\vspace{2mm}
\begin{tcolorbox}[
  colback=GEGreenLight, colframe=GEGreen, coltitle=white, fonttitle=\bfseries\small,
  title=Best Trajectory \quad R=0.875 \quad (image: 1.0 / text: 0.75),
  sharp corners, boxrule=0.5mm, left=2mm, right=2mm, top=1.5mm, bottom=1.5mm, breakable
]
\footnotesize
\textbf{Turn 1 --- Think \& Tool Call:} \texttt{search("A\'{e}rotrain I80 world record top speed year set")}\\
\textit{``\ldots first, I need to find two key things: the exact top speed of the A\'{e}rotrain I80's world record run and the year it was set, right? Because the user wants that exact text on the poster\ldots''}

\vspace{1mm}
\textbf{$\rightarrow$ Observation:} Returns speed 430.4\,km/h (267\,mph) set in \textbf{1974}.

\vspace{1mm}
\textbf{Turn 2 --- Think \& Tool Call:} \texttt{image\_search("A\'{e}rotrain I80 on inverted T concrete track")}\\
\textit{``\ldots the record speed is 430.4\,km/h set in 1974. Good. Now I need a visual reference of the A\'{e}rotrain I80 and its inverted T-shaped track to make the poster accurate\ldots''}

\vspace{1mm}
\textbf{$\rightarrow$ Observation:} Returns clear images. Selects \textbf{IMG\_001}.

\vspace{1mm}
\textbf{Turn 3 --- Think \& Tool Call:} \texttt{query\_knowledge("text\_rendering")} \quad \textcolor{GEGreen}{$\leftarrow$ critical skill}\\
\textit{``\ldots IMG\_001 is a clear shot of the A\'{e}rotrain I80 on its track. Now I need to get guidance on text rendering for 1970s vintage transport advertisement typography, right? Because the user wants bold vintage typography with the exact speed and year\ldots''}

\vspace{1mm}
\textbf{Turn 4 --- Think \& Final Answer:}\\
\textit{``\ldots the gen\_prompt should be retro-futuristic 1970s travel poster style\ldots Bold vintage 1970s transport advertisement typography centered in the lower third of the poster: first line reads `A\'{E}ROTRAIN I80' in large bold rounded retro sans-serif font, second line reads `430.4\,km/h \textbullet{} 267\,mph', third line reads `WORLD SPEED RECORD \textbullet{} 1974'\ldots''}

\vspace{1mm}
\textbf{Final program (gen\_prompt):} \textit{``Retro-futuristic 1970s-style travel poster, featuring the A\'{e}rotrain I80 gliding on its inverted T-shaped concrete track as shown in the first reference image. Bold vintage 1970s transport advertisement typography centered in the lower third of the poster: first line reads `A\'{E}ROTRAIN I80' in large, bold, rounded retro sans-serif font, second line reads `430.4\,km/h \textbullet{} 267\,mph' in slightly smaller bold matching font, third line reads `WORLD SPEED RECORD \textbullet{} 1974'\ldots Vibrant, saturated 1970s color palette, warm golden hour lighting, clean retro composition\ldots''}\\
\textbf{Skills called:} \texttt{text\_rendering} \checkmark \quad\textbf{Reference:} IMG\_001 (A\'{e}rotrain on track). \quad\textbf{Judge:} \textbf{all pass}.
\end{tcolorbox}

\vspace{2mm}
\begin{tcolorbox}[
  colback=GERedLight, colframe=GERed, coltitle=white, fonttitle=\bfseries\small,
  title=Worst Trajectory \quad R=0.40 \quad (image: 0.05 / text: 0.75),
  sharp corners, boxrule=0.5mm, left=2mm, right=2mm, top=1.5mm, bottom=1.5mm, breakable
]
\footnotesize
\textbf{Turn 1 --- Think \& Tool Call:} \texttt{search("A\'{e}rotrain I80 official world record top speed year of record run")}\\
\textit{``\ldots first I should do a text search for that, because the user wants the exact speed and year for the text\ldots''}

\vspace{1mm}
\textbf{$\rightarrow$ Observation:} Same facts: 430.4\,km/h, 1974.

\vspace{1mm}
\textbf{Turn 2 --- Think \& Tool Call:} \texttt{image\_search("A\'{e}rotrain I80 gliding on inverted T shaped concrete track")}\\
\textit{``\ldots the world record top speed is 430.4\,km/h set in 1974. Perfect. Now I need visual references: first, an image of the A\'{e}rotrain I80 on its track\ldots''}

\vspace{1mm}
\textbf{Turn 3 --- Think \& Tool Call:} \texttt{image\_search("1970s retro-futuristic travel poster vintage transport advertisement typography")}\\
\textit{``\ldots IMG\_001 is perfect, it shows the A\'{e}rotrain I80 clearly on its T-shaped track. Now I need a reference for 1970s retro-futuristic travel poster style with vintage transport typography\ldots''}

\vspace{1mm}
\textbf{Turn 4 --- Think \& Final Answer:} \quad \textcolor{GERed}{$\leftarrow$ \textbf{no} \texttt{query\_knowledge} \textbf{called at all}}\\
\textit{``\ldots I have both references I need. First, IMG\_001 for the A\'{e}rotrain. Second, IMG\_007 for the retro-futuristic travel poster style\ldots Use the 1970s vintage travel poster aesthetic, warm saturated color palette, and bold retro typography style from the second reference image\ldots''}

\vspace{1mm}
\textbf{Final program (gen\_prompt):} \textit{``Retro-futuristic 1970s-style travel poster, featuring the A\'{e}rotrain I80 gliding on its inverted T-shaped concrete track as shown in the first reference image. Use the 1970s vintage travel poster aesthetic\ldots from the second reference image. Prominent bold vintage transport advertisement text across the poster reads: `A\'{e}rotrain I80 | Official World Speed Record: 430.4\,km/h (267\,mph) | 1974'\ldots''}\\
\textbf{Skills called:} \textbf{none}. \quad \textcolor{GERed}{\textbf{Missing:} \texttt{text\_rendering} --- no guidance on typography decomposition or spatial placement of text.}

\vspace{1mm}
\textcolor{GERed}{\textbf{Critical failures:}}
\begin{enumerate}[leftmargin=1.2em, itemsep=0pt, topsep=1pt]
\item \textbf{Skipped} \texttt{text\_rendering} --- crammed all text into a single long string ``A\'{e}rotrain I80 | Official World Speed Record: 430.4\,km/h (267\,mph) | 1974'' instead of decomposing into separate lines with spatial anchors $\to$ \texttt{text\_rendering}=\textbf{fail}.
\item Used a second style reference image instead of skill guidance --- relied on the model to mimic typography from a reference rather than applying explicit font/layout rules $\to$ \texttt{aesthetic}=\textbf{fail}, \texttt{attribute}=\textbf{fail}.
\item Same correct facts as best, but without \texttt{text\_rendering} guidance the poster is unreadable: \texttt{spatial\_layout}=\textbf{fail}.
\end{enumerate}
\end{tcolorbox}

\begin{figure}[h]
\centering
\begin{minipage}{0.48\linewidth}\centering
\includegraphics[width=\linewidth]{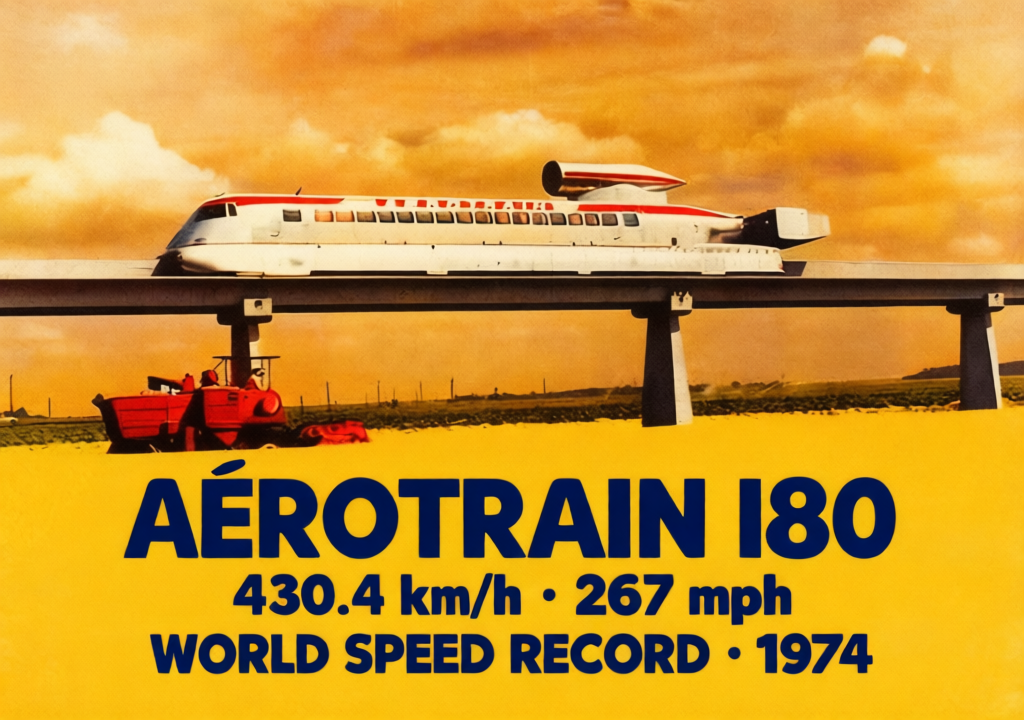}\\[2pt]
{\small\textcolor{GEGreen}{\textbf{Best} ($R=0.875$): clear typography, correct layout}}
\end{minipage}\hfill
\begin{minipage}{0.48\linewidth}\centering
\includegraphics[width=\linewidth]{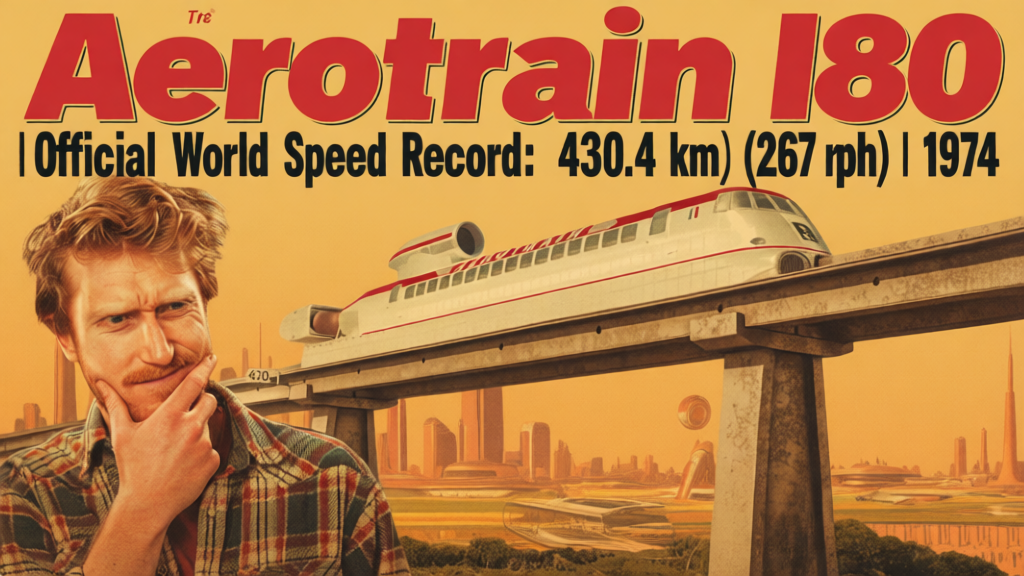}\\[2pt]
{\small\textcolor{GERed}{\textbf{Worst} ($R=0.40$): garbled text, broken poster}}
\end{minipage}
\caption{\textbf{Case 2 generated images.} Both trajectories retrieved the same correct facts (430.4\,km/h, 1974). The best trajectory called \texttt{text\_rendering} and decomposed text into explicit lines with spatial anchors. The worst skipped all skills and crammed text into one string, resulting in unreadable typography.}
\label{fig:case2_images}
\end{figure}

\vspace{2mm}
\begin{tcolorbox}[
  colback=GEAmberLight, colframe=GEAmber, coltitle=white, fonttitle=\bfseries\small,
  title=Case 2 --- Extracted Experience Slots \quad (Delta = 0.475 above threshold),
  sharp corners, boxrule=0.5mm, left=2mm, right=2mm, top=1.5mm, bottom=1.5mm
]
\footnotesize
\begin{itemize}[leftmargin=1.2em, itemsep=2pt, topsep=2pt]
\item[\textcolor{GEBlue}{\textbf{S1}}] \textbf{Search strategy:} Execute parallel searches for both the historical event (speed/year) and the physical design (inverted T-track) to ground both text and visuals.
\item[\textcolor{GEAmber}{\textbf{S2}}] \textbf{Knowledge activation:} Call \texttt{text\_rendering} to obtain guidance on line-level text decomposition, spatial placement, and typographic styling --- a style reference image alone cannot replace this.
\item[\textcolor{GECyan}{\textbf{S3}}] \textbf{Reference selection:} Select one reference for the vehicle/infrastructure accuracy; do not add a second ``style'' reference as a substitute for skill guidance.
\item[\textcolor{GEGreen}{\textbf{S4}}] \textbf{Prompt construction:} Decompose text into explicit separate lines (e.g.\ first line, second line) and assign a specific spatial location (e.g.\ ``lower third'') instead of a single concatenated string.
\item[\textcolor{GERed}{\textbf{S5}}] \textbf{Failure avoidance:} Cramming multiple text elements into one line with separators (``|'') causes the generator to render garbled, overlapping text --- always decompose into spatially anchored lines.
\end{itemize}
\vspace{1mm}
% {\scriptsize\textcolor{GEMuted}{Case 2: Same facts, same subject reference. The difference is purely whether \texttt{text\_rendering} skill was called: best decomposed text into 3 styled lines; worst dumped it as one string.}}
\end{tcolorbox}

%% ═══════════════════════ CASE 3: European Housing ═══════════════════════

\vspace{4mm}
\begin{tcolorbox}[
  colback=GELight, colframe=GEMuted, coltitle=white, fonttitle=\bfseries\small,
  title=Case 3 --- User Request, sharp corners, boxrule=0.5mm,
  left=2mm, right=2mm, top=1.5mm, bottom=1.5mm
]
\small
\textit{``Generate a street view with two famous European housing complexes side by side. On the left, feature the colorful, uneven apartment building designed by Friedensreich Hundertwasser in Vienna. On the right, feature the tilted yellow houses designed by Piet Blom in Rotterdam. Above the entrance of the left building, add a large wooden sign reading `ORGANIC'. Above the right building, place a sleek metal sign reading `GEOMETRIC'.''}
\end{tcolorbox}

\vspace{2mm}
\begin{tcolorbox}[
  colback=GEGreenLight, colframe=GEGreen, coltitle=white, fonttitle=\bfseries\small,
  title=Best Trajectory \quad R=0.80 \quad (image: 0.6 / text: 1.0),
  sharp corners, boxrule=0.5mm, left=2mm, right=2mm, top=1.5mm, bottom=1.5mm, breakable
]
\footnotesize
\textbf{Turn 1 --- Think \& Tool Call:} \texttt{image\_search("Friedensreich Hundertwasser Vienna colorful uneven apartment building exterior")}\\
\textit{``\ldots first, I need to get visual references for both housing complexes. Let's start with an image search for Hundertwasser building Vienna\ldots''}

\vspace{1mm}
\textbf{Turn 2 --- Think \& Tool Call:} \texttt{image\_search("Piet Blom Rotterdam tilted yellow houses exterior")}\\
\textit{``\ldots IMG\_001 is the clear exterior shot, that's perfect for the left building. Now I need the reference for the Piet Blom Rotterdam tilted yellow houses\ldots''}

\vspace{1mm}
\textbf{Turn 3 --- Think \& Tool Call:} \texttt{query\_knowledge("spatial\_layout")} \quad \textcolor{GEGreen}{$\leftarrow$ critical skill}\\
\textit{``\ldots now I have both building references. First, spatial layout: because we have left/right arrangement of buildings, signs above each entrance, street view perspective\ldots''}

\vspace{1mm}
\textbf{Turn 4 --- Think \& Tool Call:} \texttt{query\_knowledge("text\_rendering")}\\
\textit{``\ldots I need text rendering guidance because there are two signs with specific text, different materials\ldots''}

\vspace{1mm}
\textbf{Turn 5 --- Think \& Tool Call:} \texttt{query\_knowledge("physical\_material\_consistency")}\\
\textit{``\ldots I need physical material consistency guidance to make sure the building materials, sign materials are correct\ldots''}

\vspace{1mm}
\textbf{Turn 6 --- Think \& Final Answer:}\\
\textit{``\ldots spatial layout: foreground is paved grey asphalt street with a narrow concrete sidewalk running between the two buildings\ldots Midground left side of the frame: the colorful uneven Hundertwasser apartment building\ldots Midground right side: the tilted yellow cube houses\ldots Centered directly above the main entrance of this left building: a large rough natural wood sign\ldots reading `ORGANIC'\ldots''}

\vspace{1mm}
\textbf{Final program (gen\_prompt):} \textit{``Realistic street view photograph, soft overcast daylight\ldots Foreground: paved grey asphalt street with a narrow concrete sidewalk running between the two buildings\ldots Midground left side of the frame: the colorful uneven Hundertwasser apartment building, fully matching the architectural details\ldots of the first reference image\ldots Centered directly above the main entrance of this left building: a large rough natural wood sign, with bold black sans-serif text carved into the wood reading `ORGANIC'\ldots Midground right side\ldots the tilted yellow cube houses\ldots a sleek brushed steel sign\ldots reading `GEOMETRIC'\ldots Both buildings are at equal height, spaced 10 feet apart with the sidewalk between them\ldots''}\\
\textbf{Skills called:} \texttt{spatial\_layout} \checkmark, \texttt{text\_rendering} \checkmark, \texttt{physical\_material\_consistency} \checkmark\\
\textbf{Judge:} \texttt{spatial}=\textbf{pass}, \texttt{text}=\textbf{pass}, \texttt{attribute}=\textbf{pass}, \texttt{material}=\textbf{pass}.
\end{tcolorbox}

\vspace{2mm}
\begin{tcolorbox}[
  colback=GERedLight, colframe=GERed, coltitle=white, fonttitle=\bfseries\small,
  title=Worst Trajectory \quad R=0.40 \quad (image: 0.05 / text: 0.75),
  sharp corners, boxrule=0.5mm, left=2mm, right=2mm, top=1.5mm, bottom=1.5mm, breakable
]
\footnotesize
\textbf{Turn 1 --- Think \& Tool Call:} \texttt{image\_search("Friedensreich Hundertwasser Vienna colorful uneven apartment building exterior")}\\
\textit{``\ldots first, I need to get visual references for both housing complexes. Let's start with image search for Hundertwasser Vienna building\ldots''}

\vspace{1mm}
\textbf{Turn 2 --- Think \& Tool Call:} \texttt{image\_search("Piet Blom Rotterdam tilted yellow houses exterior street view")}\\
\textit{``\ldots IMG\_003 looks like a clear, standard exterior shot. Now I need to get the reference for the Piet Blom Rotterdam tilted yellow houses\ldots''}

\vspace{1mm}
\textbf{Turn 3 --- Think \& Tool Call:} \texttt{query\_knowledge("text\_rendering")}\\
\textit{``\ldots now I have both building references. Next, I need guidance for text rendering because there are two specific signs: wooden `ORGANIC' above the left building, metal `GEOMETRIC' above the right\ldots''}

\vspace{1mm}
\textbf{Turn 4 --- Think \& Final Answer:} \quad \textcolor{GERed}{$\leftarrow$ skipped \texttt{spatial\_layout} and \texttt{physical\_material}}\\
\textit{``\ldots I have all the info I need\ldots the gen\_prompt needs to follow the text rendering guidance, plus spatial layout: left is Hundertwasser building, right is cube houses side by side\ldots''}

\vspace{1mm}
\textbf{Final program (gen\_prompt):} \textit{``Photorealistic sunny daytime European street view, two famous housing complexes placed side by side at equal width. On the left, copy the colorful uneven facade\ldots from the first reference image. On the right, copy the tilted geometric shape\ldots from the second reference image. Above the main entrance of the left building, place a large weathered wooden sign\ldots spelling out `ORGANIC'\ldots Above the main entrance of the right building, place a sleek brushed metal sign\ldots spelling out `GEOMETRIC'\ldots Foreground has a cobblestone street\ldots''}\\
\textbf{Skills called:} \texttt{text\_rendering} \checkmark \quad \textcolor{GERed}{\textbf{Missing:} \texttt{spatial\_layout}, \texttt{physical\_material\_consistency}}

\vspace{1mm}
\textcolor{GERed}{\textbf{Critical failures:}}
\begin{enumerate}[leftmargin=1.2em, itemsep=0pt, topsep=1pt]
\item \textbf{Missing} \texttt{spatial\_layout} --- used vague ``placed side by side at equal width'' without frame-relative coordinates $\to$ buildings overlap or merge, \texttt{spatial\_layout}=\textbf{fail}.
\item Without precise spatial anchors, text signs float or attach to the wrong building $\to$ \texttt{text\_rendering}=\textbf{fail}, \texttt{attribute\_binding}=\textbf{fail}.
\item \textbf{Missing} \texttt{physical\_material\_consistency} --- sign materials (wood vs metal) not properly grounded $\to$ \texttt{physical\_material}=partial.
\end{enumerate}
\end{tcolorbox}

\begin{figure}[h]
\centering
\begin{minipage}{0.48\linewidth}\centering
\includegraphics[width=\linewidth]{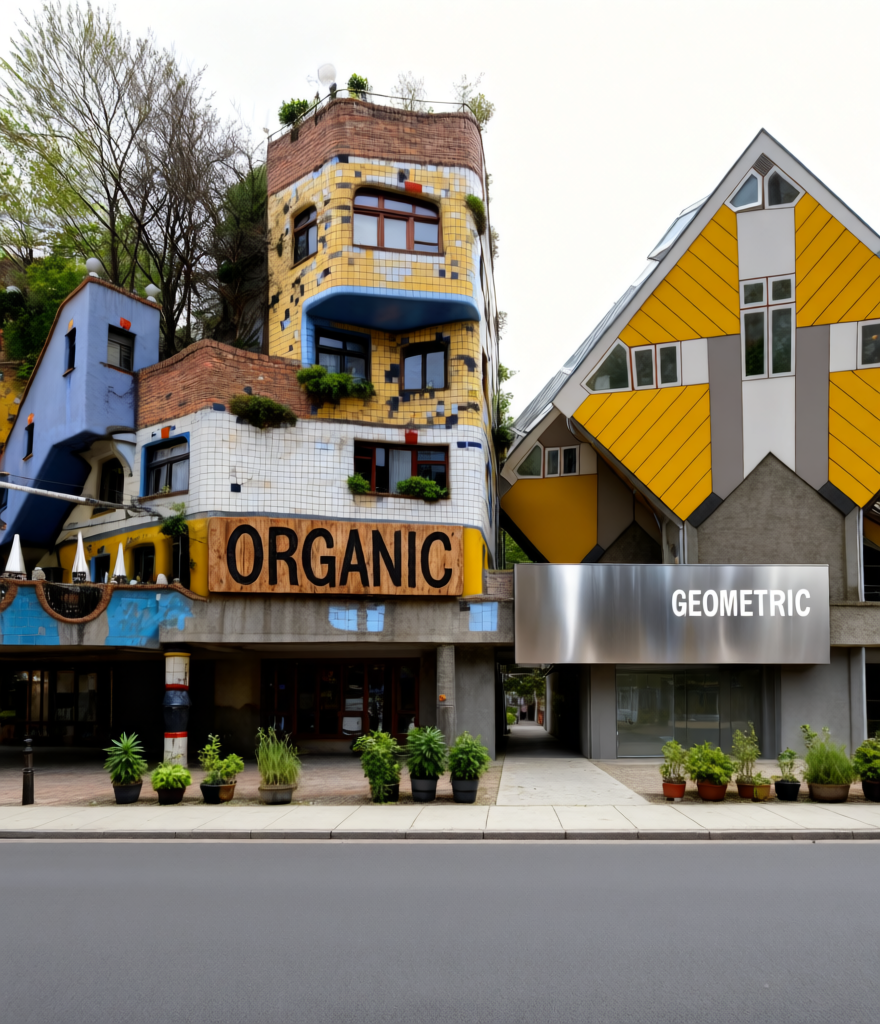}\\[2pt]
{\small\textcolor{GEGreen}{\textbf{Best} ($R=0.80$): correct layout, both signs legible}}
\end{minipage}\hfill
\begin{minipage}{0.48\linewidth}\centering
\includegraphics[width=\linewidth]{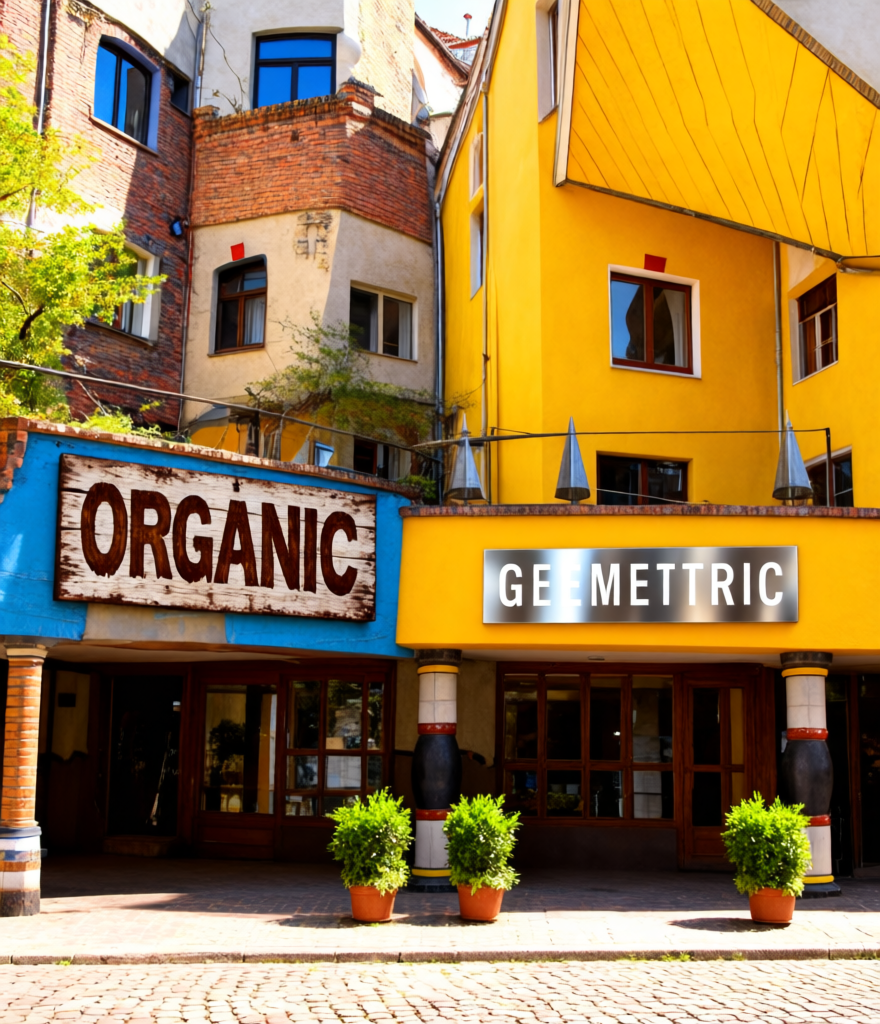}\\[2pt]
{\small\textcolor{GERed}{\textbf{Worst} ($R=0.40$): merged buildings, text failure}}
\end{minipage}
\caption{\textbf{Case 3 generated images.} The best trajectory called \texttt{spatial\_layout} and used frame-relative coordinates (``midground left/right side of the frame, spaced 10 feet apart''). The worst skipped \texttt{spatial\_layout} and used vague ``side by side at equal width,'' causing the buildings to merge and text signs to fail.}
\label{fig:case3_images}
\end{figure}

\vspace{2mm}
\begin{tcolorbox}[
  colback=GEAmberLight, colframe=GEAmber, coltitle=white, fonttitle=\bfseries\small,
  title=Case 3 --- Extracted Experience Slots \quad (Delta = 0.475 above threshold),
  sharp corners, boxrule=0.5mm, left=2mm, right=2mm, top=1.5mm, bottom=1.5mm
]
\footnotesize
\begin{itemize}[leftmargin=1.2em, itemsep=2pt, topsep=2pt]
\item[\textcolor{GEBlue}{\textbf{S1}}] \textbf{Search strategy:} Include the requested camera angle in all image search queries to ensure consistent perspective references for the street view.
\item[\textcolor{GEAmber}{\textbf{S2}}] \textbf{Knowledge activation:} Call \texttt{spatial\_layout} to obtain guidance on using absolute frame-relative coordinates --- \texttt{text\_rendering} alone cannot prevent spatial overlap or attribute leakage between adjacent buildings.
\item[\textcolor{GECyan}{\textbf{S3}}] \textbf{Reference selection:} Explicitly state the intended spatial position (e.g.\ ``placed on the left side of the scene'') in the reference image notes.
\item[\textcolor{GEGreen}{\textbf{S4}}] \textbf{Prompt construction:} Use absolute frame-relative coordinates (e.g.\ ``Midground left side'') and explicitly mount text signs to these specific spatial zones with depth layering.
\item[\textcolor{GERed}{\textbf{S5}}] \textbf{Failure avoidance:} Strictly partition the prompt using depth layers and absolute horizontal positions to isolate the architectural and textual attributes --- vague ``side by side'' causes attribute bleeding between buildings.
\end{itemize}
\vspace{1mm}
% {\scriptsize\textcolor{GEMuted}{Case 3: Both trajectories searched the same buildings and called \texttt{text\_rendering}. The difference is purely the missing \texttt{spatial\_layout}: without it, the generator could not separate two distinct buildings and their respective text signs.}}
\end{tcolorbox}

\subsection{Prompt-Keyed Visual Experience Memory}

For each prompt, the self-evolution pipeline samples multiple rollouts. After image generation and judging, the system identifies the best and worst rollout by scalar reward. If the reward gap exceeds a minimum margin, the comparison is added to a pending buffer. Comparisons caused only by missing references or other protocol failures are ignored because they do not provide a reusable visual strategy.

At each memory update, high-gap comparisons are summarized into five dimensions: search strategy,
knowledge activation, reference selection, prompt construction, and failure avoidance. The memory is prompt-keyed rather than global. Each entry is attached to the source prompt that produced the best-worst comparison and stores a source-prompt embedding matched to the generated strategy key and other information. Each slot keeps a capacity-limited buffer and trims
by reward gap and recency when full.

\begin{table}[h]
\centering
\footnotesize
\caption{Prompt-keyed visual experience slots used to construct the privileged teacher context.}
\label{tab:appendix_experience_slots}
\setlength{\tabcolsep}{4pt}
\renewcommand{\arraystretch}{1.12}
\resizebox{\linewidth}{!}{%
\begin{tabular}{p{0.20\linewidth} p{0.34\linewidth} p{0.36\linewidth}}
\toprule
\rowcolor{GEGreen!10}
Slot & Extracted contrast & Teacher-side effect \\
\midrule
\smallpill{GEBlue}{S1} \slot{search strategy} 
& Factual gap, query decomposition, evidence verification, and stopping condition 
& Guides whether to search, what to verify, and when evidence is sufficient \\

\rowcolor{GELight!55}
\smallpill{GEAmber}{S2} \slot{knowledge activation} 
& Called, missed, or unnecessary generation knowledge explaining the reward gap 
& Biases \tool{query\_knowledge} calls toward relevant visible failure modes \\

\smallpill{GECyan}{S3} \slot{reference selection} 
& Useful, redundant, or harmful retrieved image candidates and their intended roles 
& Improves filtering, deduplication, and ordinal reference binding \\

\rowcolor{GELight!55}
\smallpill{GEGreen}{S4} \slot{prompt construction} 
& Program structure that carries evidence, reference roles, and knowledge guidance into generation 
& Improves prompt-reference program synthesis \\

\smallpill{GERed}{S5} \slot{failure avoidance} 
& Recurring low-reward visual patterns for similar requests 
& Adds guards for text, counting, material, anatomy, grounding, and style failures \\
\bottomrule
\end{tabular}}
\end{table}

\subsection{Source-Prompt Bundle Retrieval}
\label{subsec:source_prompt_retrieval}

When constructing teacher context for a current prompt, the system retrieves by source-prompt similarity rather than by experience-text similarity. Let \(\mathcal{B}\) be the set of historical source prompts stored in the memory. The retrieved source prompt is
\begin{equation}
  \tilde{x}=\arg\max_{x_j\in\mathcal{B}}\cos(e(x),e(x_j)).
\end{equation}
The teacher receives the entries from all five slots that share the same source prompt \(\tilde{x}\). This bundle retrieval avoids mixing unrelated lessons from different cases. If no entry exists, no teacher context is produced and the SDL term is skipped for that row.

\subsection{GRPO Rollout Loss}
\label{subsec:grpo_loss}

For each prompt \(x\), the behavior policy samples \(K\) complete visual rollouts. Each rollout contains tool-call tokens, final-answer tokens, selected references, the generated image, and a scalar reward. Environment observations and generated images are not optimized directly; the policy loss is computed only on assistant tokens with mask \(m_{i,t}\).

Let \(R_i\) be the final mixed reward of rollout \(i\). The group-relative advantage is
\begin{equation}
  \widehat{A}_i=\frac{R_i-\bar{R}}{\sigma_R+\epsilon_{\mathrm{adv}}},
  \qquad
  \bar{R}=\frac{1}{K}\sum_{j=1}^{K}R_j,\quad
  \sigma_R=\sqrt{\frac{1}{K}\sum_{j=1}^{K}(R_j-\bar{R})^2}.
\end{equation}
For a sampled token \(y_{i,t}\), the policy ratio is
\begin{equation}
  u_{i,t}(\theta)=
  \exp\!\left(
  \log\pi_\theta(y_{i,t}\mid h_{i,t})
  -\operatorname{sg}\!\left(\log\pi_{\theta_{\mathrm{old}}}(y_{i,t}\mid h_{i,t})\right)
  \right).
\end{equation}
The clipped GRPO surrogate minimized by the agent is
\begin{equation}
\mathcal{L}_{\mathrm{GRPO}}=
-\frac{1}{\sum_{i,t}m_{i,t}}
\sum_{i,t}m_{i,t}
\min\!\left(
u_{i,t}\widehat{A}_i,
\mathrm{clip}(u_{i,t},1-\epsilon_{\ell},1+\epsilon_h)\widehat{A}_i
\right)
+\beta_{\mathrm{ref}}\mathcal{K}_{\mathrm{ref}},
\end{equation}
where \(\mathcal{K}_{\mathrm{ref}}\) denotes the reference-policy regularization term when enabled. The same final reward supplies the group advantage for all optimized assistant tokens in a rollout, including tool decisions and final prompt-reference program construction.

\subsection{Experience-Conditioned SDL Contexts}
\label{subsec:sdl_details}

Experience-conditioned SDL uses two contexts for the same sampled assistant tokens. The student receives the normal inference context. The teacher receives the same context with the retrieved source-prompt experience bundle inserted into the system prompt before the tool definitions. Teacher and student share model weights; the teacher view is detached and privileged only during training. The returned policy is the student policy and does not require dynamic experience slots at inference.

The sampled-token reverse-KL estimator and the importance correction follow prior self-distillation and KL-estimator work~\citep{skillsd,schulman2020kl,tang2025kl,liu2025kl}; the GenEvolve-specific component is the construction and retrieval of visual experience for the teacher branch. For token \(y_{i,t}\) in rollout \(i\), define
\begin{equation}
p_{i,t}=\pi^{S}_{\theta}(y_{i,t}\mid h_{i,t}),\qquad
q_{i,t}=\operatorname{sg}\!\left[\pi^{E}_{\theta}(y_{i,t}\mid \tilde{h}_{i,t})\right],\qquad
\ell_{i,t}=\log p_{i,t}-\log q_{i,t}.
\end{equation}
The sampled-token KL estimator is
\begin{equation}
  k_3(\ell_{i,t})=\exp(-\ell_{i,t})-1+\ell_{i,t}.
\end{equation}
Let \(m^{E}_{i,t}\) select valid assistant tokens from rows with real teacher context, and let
\begin{equation}
  \rho^{\mathrm{on}}_{i,t}=
  \min\!\left(
  \frac{\pi^{S}_{\theta}(y_{i,t}\mid h_{i,t})}
  {\operatorname{sg}\!\left[\pi^{S}_{\theta_{\mathrm{old}}}(y_{i,t}\mid h_{i,t})\right]},
  \rho_{\max}
  \right)
\end{equation}
be the clipped student-centered on-policy importance ratio. The implemented SDL loss is
\begin{equation}
\mathcal{L}_{\mathrm{SDL}}=
\frac{1}{\sum_{i,t}m^E_{i,t}}
\sum_{i,t}m^E_{i,t}
\min\!\left(\rho^{\mathrm{on}}_{i,t}k_3(\ell_{i,t}),c_{\mathrm{tok}}\right).
\end{equation}
SDL is applied on the same on-policy responses used by the group-relative rollout loss, so it adds dense token-level guidance without introducing a separate offline imitation dataset. Compared with skill-conditioned self-distillation for text agents, the teacher conditioning object here is a prompt-keyed visual experience bundle extracted from generated-image best-worst trajectory pairs. The SDL coefficient and clipping constants are kept fixed across all reported ablations.

\begin{figure}[t]
    \centering
    \includegraphics[width=\linewidth]{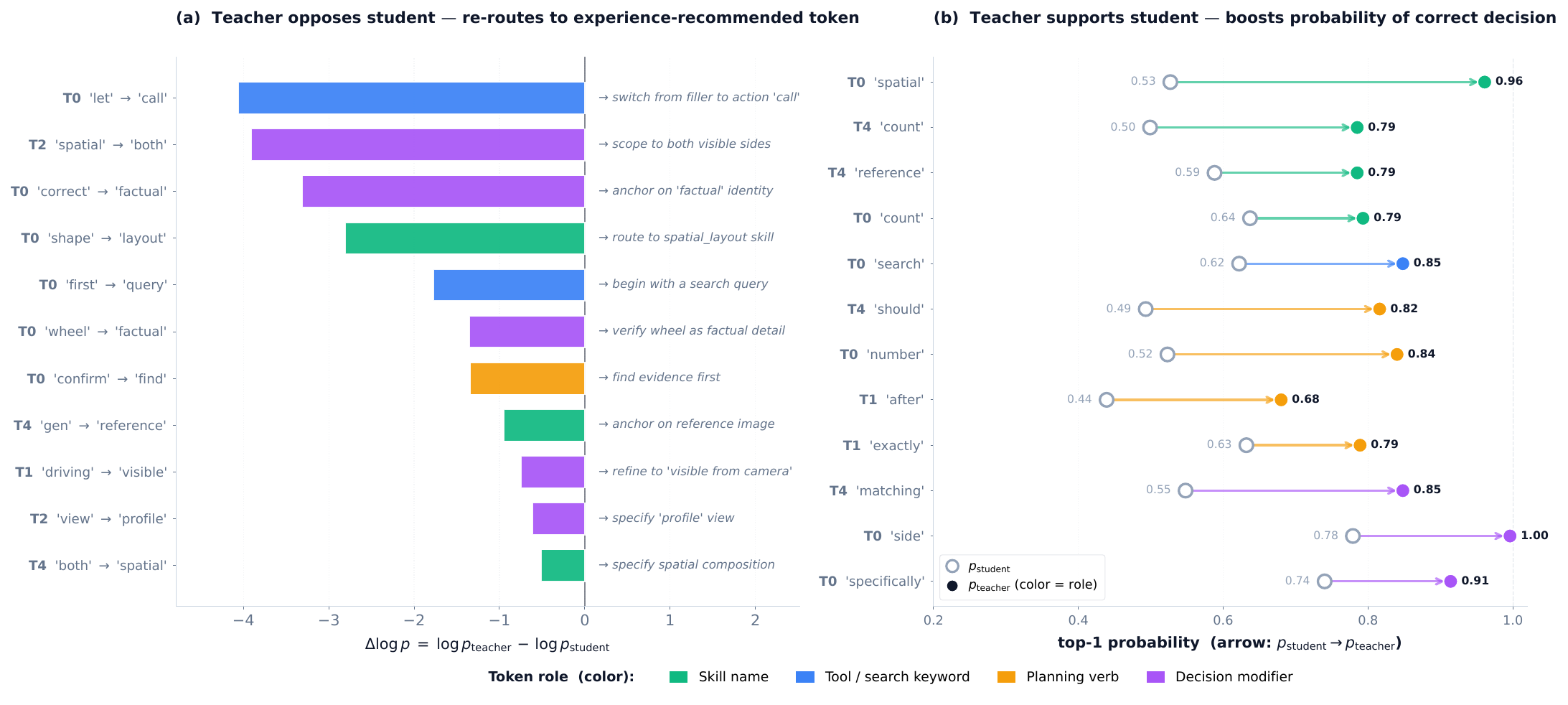}
    \caption{\textbf{Token-level evidence of experience-conditioned SDL guidance.}
    Representative tokens from a single held-out rollout illustrate the two complementary effects of the teacher signal under the prompt-keyed experience bundle. The case asks for a stylised rendering of the \emph{Wuppertal Schwebebahn} that must respect a real landmark's identity, layout and a specified visible-carriage count; the bundle instructs the agent to verify the count from a specified perspective using image references before invoking quantity counting.
    \textbf{(a)} Teacher \emph{opposes} the student: at decision tokens where the student commits to a generic or off-target word (e.g., ``shape'', ``correct'', ``first'', ``gen''), the teacher concentrates its mass on an experience-recommended alternative (e.g., ``layout'', ``factual'', ``query'', ``reference''), producing a large negative \(\Delta\log p\). Bar color encodes the role of the teacher-recommended token: \emph{skill} name, \emph{tool}/search keyword, planning \emph{verb}, or decision \emph{modifier}.
    \textbf{(b)} Teacher \emph{supports} the student: at tokens where the student is already on-policy but uncertain, the teacher boosts the same top-1 token from \(p_{\mathrm{student}}\) (hollow circle) to \(p_{\mathrm{teacher}}\) (filled circle, color = role). Both effects appear at multiple turns (T0--T4) and consistently track the case-specific bundle (factual identity, reference-based counting, spatial composition), confirming that SDL provides dense token-level guidance complementary to the trajectory-level GRPO reward.}
    \label{fig:sdl_token_evidence}
\end{figure}

\paragraph{Token-level evidence of teacher guidance.}
\label{para:sdl_token_evidence}
To verify that experience-conditioned SDL indeed provides actionable token-level guidance rather than merely matching the student distribution, we instrument a representative held-out rollout and inspect the teacher-vs-student log-probabilities at the assistant tokens where the SDL loss is applied. We use a request to render the \emph{Wuppertal Schwebebahn} (a historical suspended monorail in Germany) in a stylised illustration that simultaneously requires (i)~factual visual identity of a real-world landmark, (ii)~the exact number of visible carriages from a specified viewpoint, and (iii)~a non-photorealistic style/layout transfer. The retrieved experience bundle for this prompt focuses on \emph{verifying the exact count of visible components from the specified perspective using image references before applying quantity counting}, which couples three skills the agent must coordinate: factual search, visual reference selection, and quantity counting. Figure~\ref{fig:sdl_token_evidence} shows two complementary effects of the teacher signal on this trajectory:

\begin{itemize}[leftmargin=1.2em, itemsep=2pt, topsep=2pt]
\item \textbf{(a) Teacher opposes student.}
On tokens where the student commits to a less effective lexical choice, the teacher concentrates its mass on a different, experience-recommended token, producing a strongly negative \(\Delta\log p=\log p_{\mathrm{teacher}}-\log p_{\mathrm{student}}\). The opposed tokens span multiple turns (T0, T1, T2, T4) and three decision aspects of this case. \emph{(i) Search vs free reasoning at task entry:} at T0 the teacher rewrites planning fillers into explicit tool calls (``let''~\(\to\)~``call'', ``first''~\(\to\)~``query'', ``confirm''~\(\to\)~``find''), enforcing the bundle's instruction to gather factual evidence before describing the scene. \emph{(ii) Skill activation:} at T0 the teacher replaces the generic descriptor ``shape'' with the named skill ``layout'' (routing the agent into \texttt{spatial\_layout}), and at T4 it pushes ``gen''~\(\to\)~``reference'' and ``both''~\(\to\)~``spatial'' when the student is composing the final program, anchoring the description on the retrieved reference and on spatial composition. \emph{(iii) Grounded modifiers:} the teacher tightens loose adjectives into the bundle's grounding cues (``correct''~\(\to\)~``factual'', ``wheel''~\(\to\)~``factual'', ``view''~\(\to\)~``profile'', ``driving''~\(\to\)~``visible''), forcing the agent to express the condition in the specific visual form the bundle prescribes.
\item \textbf{(b) Teacher supports student.}
On tokens where the student is already moving in the correct direction but uncertain, the teacher \emph{boosts} the same top-1 token's probability, producing a positive gap \(p_{\mathrm{teacher}}-p_{\mathrm{student}}>0\). The boosts coincide with the same case-specific concepts: at T0 the teacher reinforces the routing into the \texttt{spatial\_layout} and \texttt{quantity\_counting} skills (\texttt{`spatial'}: 0.527\(\to\)0.961, \texttt{`count'}: 0.637\(\to\)0.793) and the choice of search as the first action (\texttt{`search'}: 0.622\(\to\)0.848); at T4 it strengthens the binding to the retrieved image (\texttt{`reference'}: 0.588\(\to\)0.785) and the explicit count statement in the final program (\texttt{`count'}: 0.499\(\to\)0.785). The reinforced tokens cover skill names, search keywords, planning verbs, and decision modifiers, matching the experience-bundle pattern observed in (a).
\end{itemize}

The two patterns together show that experience-conditioned SDL acts as a fine-grained controller: it re-routes the agent at the few decision tokens where free-form generation would diverge from the experience-distilled policy, while simultaneously sharpening the agent's confidence on the many tokens where the student is already correct. Both effects in this case track the same underlying experience bundle (factual identity, reference-based counting, spatial composition), confirming that the SDL signal is content-specific rather than a generic regulariser, and that it complements the trajectory-level GRPO reward with dense token-level guidance.

\section{Training Details}

\subsection{Supervised Stage Configuration}

The supervised stage cold-starts Qwen3-VL-8B-Instruct into a tool-orchestrated image-generation agent.
We fine-tune the language-policy part on supervised GenEvolve-Data trajectories using a long-context multimodal training stack, while keeping the visual encoder fixed.
Only assistant-side tokens are optimized, including reasoning, tool calls, and final prompt-reference programs; user prompts and tool observations are used as context and masked from the loss.
Concrete hyper-parameters, optimizer settings, batch sizes, and the LLaMA-Factory configuration we use are summarised in Table~\ref{tab:appendix_sft}.

\begin{table}[h]
\centering
\footnotesize
\caption{Supervised trajectory-tuning configuration. We list the concrete values used for the released checkpoint; the same setup is used for all SFT-based baselines reported in the main paper.}
\label{tab:appendix_sft}
\setlength{\tabcolsep}{4pt}
\renewcommand{\arraystretch}{1.12}
\resizebox{\linewidth}{!}{%
\begin{tabular}{p{0.30\linewidth} p{0.36\linewidth} p{0.32\linewidth}}
\toprule
\rowcolor{GEBlue!10}
Setting & Value & Notes \\
\midrule
Backbone 
& Qwen3-VL-8B-Instruct~\citep{qwen3vl}
& Base multimodal agent policy. \\

\rowcolor{GELight!55}
Framework 
& LLaMA-Factory~\citep{llamafactory} 
& Long-context multimodal SFT with DeepSpeed ZeRO-3 sharding. \\

Trainable part 
& Language-policy parameters 
& \texttt{freeze\_vision\_tower=true}, \texttt{freeze\_multi\_modal\_projector=true}; visual encoder fixed. \\

\rowcolor{GELight!55}
Data split 
& GenEvolve-Data SFT split 
& 8{,}800 training and 200 held-out trajectories. \\

Sequence length 
& \texttt{cutoff\_len}=32{,}768 
& Long-context to fit multi-turn tool trajectories. \\

\rowcolor{GELight!55}
Optimizer 
& AdamW (default in LLaMA-Factory) 
& \texttt{weight\_decay}=$1{\times}10^{-6}$. \\

Learning rate 
& $1{\times}10^{-5}$ 
& \texttt{cosine} schedule with \texttt{warmup\_ratio}=0.02. \\

\rowcolor{GELight!55}
Epochs / batch 
& 2 epochs, 16 GPUs, micro-bsz=2, grad-accum=1 
& Effective global batch size = 32 sequences/step. \\

Precision / kernel 
& bf16, FlashAttention-2 
& Activation checkpointing on. \\

\rowcolor{GELight!55}
Loss masking 
& Assistant-token only 
& User prompts and tool-response tokens are masked from the SFT loss. \\

\bottomrule
\end{tabular}}
\end{table}

\subsection{Self-Evolution and SDL Configuration}

Self-evolution starts from the supervised trajectory checkpoint.
For each update, the agent samples multiple on-policy rollouts per prompt, and each rollout produces a prompt-reference program whose image is rendered by the reference-conditioned generator.
The generated images and final programs are scored by image and text judges, respectively (see Appendix~\ref{subsec:reward_rubric} and Table~\ref{tab:eval_rubric} for the judge backbone and rubric).
Best-worst trajectory pairs with sufficient reward gaps are mined into prompt-keyed visual experience, which is retrieved only for the privileged teacher branch during SDL.
The deployed policy is always the student branch and does not use runtime visual-experience memory.
The full set of GRPO, SDL, reward, and visual-experience-memory hyper-parameters used to train the released GenEvolve checkpoint is given in Table~\ref{tab:appendix_rl}.
Concretely, SDL is restricted to special tokens, and within each sequence we keep only the top 10\% of those tokens by $|\log\pi^E_\theta-\log\pi^S_\theta|$ (\texttt{SDL\_TOP\_K\_FRAC=0.1}); together with the importance-ratio cap $\rho_{\max}=2$ this isolates the few decision tokens where the experience-conditioned teacher disagrees most strongly with the student.

\begin{table}[!htbp]
\centering
\footnotesize
\caption{Self-evolution (GRPO + experience-conditioned SDL) configuration. We list the concrete values used to train the released GenEvolve checkpoint; all ablations in the main paper share the same setup unless stated otherwise.}
\label{tab:appendix_rl}
\setlength{\tabcolsep}{4pt}
\renewcommand{\arraystretch}{1.12}
\resizebox{\linewidth}{!}{%
\begin{tabular}{p{0.30\linewidth} p{0.36\linewidth} p{0.32\linewidth}}
\toprule
\rowcolor{GEGreen!10}
Setting & Value & Notes \\
\midrule
\multicolumn{3}{l}{\textit{\textbf{Framework and infrastructure}}} \\
RL framework 
& \texttt{rLLM}/verl 
& FSDP actor with SGLang rollout backend. \\

\rowcolor{GELight!55}
Hardware 
& 1 node $\times$ 8 GPUs 
& \texttt{fsdp\_size=8}, parameter+optimizer offload enabled. \\

Initialization 
& SFT checkpoint (Table~\ref{tab:appendix_sft}) 
& Cold start from the supervised trajectory checkpoint. \\

\midrule
\multicolumn{3}{l}{\textit{\textbf{Rollout sampling}}} \\
\rowcolor{GELight!55}
Prompt batch / rollouts 
& 8 prompts/step $\times$ 6 rollouts each
& \texttt{train\_batch\_size}=8, \texttt{n}=6 (group size for GRPO advantage). \\

Sampling 
& temperature=0.7, top-p=0.95, top-k=$-1$ 
& Same val sampling. \\

\rowcolor{GELight!55}
Max prompt / response 
& 6{,}144 / 30{,}000 tokens 
& Multi-turn tool-orchestrated rollouts. \\

Tool-call budget 
& \texttt{MAX\_LLM\_CALL\_PER\_RUN}=11 
& Allows search + image-search + skill queries plus final answer. \\

\rowcolor{GELight!55}
Generator 
& Qwen-Image-Edit-2511 (open) / Nano Banana Pro (strong) 
& References capped at \texttt{QWEN\_EDIT\_MAX\_REF\_IMAGES}=2. \\

\midrule
\multicolumn{3}{l}{\textit{\textbf{Reward (mixed image + text)}}} \\
Image judge 
& Gemini~3.1~Pro Preview, KScore protocol 
& Faithfulness/Visual/Text/Aesthetic with weights $0.1{:}0.4{:}0.4{:}0.1$. \\

\rowcolor{GELight!55}
Text judge 
& Gemini~3.1~Pro Preview, program-sufficiency 
& 5-bin scoring on $\{0,0.25,0.5,0.75,1\}$. \\

Final reward 
& $R=0.5\,R_\text{img}+0.5\,R_\text{text}$ 
& \texttt{GEN\_REWARD\_TEXT\_COEF}=0.5. \\

\midrule
\multicolumn{3}{l}{\textit{\textbf{GRPO objective}}} \\
\rowcolor{GELight!55}
Algorithm 
& Group-relative PPO surrogate 
& \texttt{adv\_estimator=grpo}, low-variance KL. \\

Learning rate 
& $1{\times}10^{-6}$ (actor) 
& Cosine, no warmup. \\

\rowcolor{GELight!55}
Clip ratios 
& $\epsilon_{\ell}=0.20$, $\epsilon_{h}=0.28$ 
& Asymmetric high clip~\citep{deepseekmath}. \\

KL regularizer 
& $\beta_\text{kl}=10^{-3}$ (KL controller) 
& \texttt{use\_kl\_loss=False}, \texttt{kl\_coef}=1e-3. \\

\rowcolor{GELight!55}
Aggregation 
& \texttt{seq-mean-token-sum} 
& Same aggregation for SDL, following Gen-Searcher~\citep{gensearcher}. \\

\midrule
\multicolumn{3}{l}{\textit{\textbf{SDL (Visual Experience Distillation)}}} \\
SDL coefficient
& $\lambda_\text{SDL}=2.0$
& \texttt{actor.sdl\_loss\_coef}=2.0. \\

\rowcolor{GELight!55}
Importance-ratio cap
& $\rho_{\max}=2.0$
& \texttt{sdl\_is\_clip}=2.0; on-policy student-centered ratio capped per token, low-variance KL estimator $k_3$ clamped to $[-10,10]$. The cap rarely fires (\texttt{sdl\_rho\_clip\_frac}$\approx 0$). \\

\midrule
\multicolumn{3}{l}{\textit{\textbf{Visual experience memory}}} \\
\rowcolor{GELight!55}
Bundle summarizer
& Gemini~3.1~Pro Preview
& \texttt{temperature}=0.0, \texttt{max\_tokens}=8192, \texttt{timeout}=90\,s, RPM cap 80. \\

Min reward gap
& $\delta_{\min}=0.20$
& \texttt{EXPERIENCE\_MIN\_REWARD\_GAP}=0.20; best/worst pair retained only if $|\Delta R|\ge\delta_\text{min}$. \\

\rowcolor{GELight!55}
Comparisons / step
& Up to 8 pairs (one per prompt group)
& \texttt{EXPERIENCE\_MAX\_COMPARISONS}=8, \texttt{TOP\_GROUPS\_PER\_STEP}=8. \\

Bundle schema
& 1 bundle per comparison
& Each bundle stores \texttt{retrieval\_key=\{trigger, source\_prompt\_summary\}} plus \texttt{decision\_guidance} (focus + 6 imperative bullet lists). \\

\rowcolor{GELight!55}
Buffer capacity
& 500 bundles, FIFO + reward-gap eviction
& \texttt{EXPERIENCE\_BUFFER\_CAPACITY}=500. \\

Prompt embedder
& Qwen3-Embedding-0.6B (CPU)
& \texttt{max\_length}=512, last-token pool with L2 normalization, cosine similarity for nearest-bundle retrieval. \\

\bottomrule
\end{tabular}}
\end{table}

\subsection{Self-Evolution Training Dynamics}
\label{subsec:training_dynamics}

\begin{figure}[h]
    \centering
    \includegraphics[width=\linewidth]{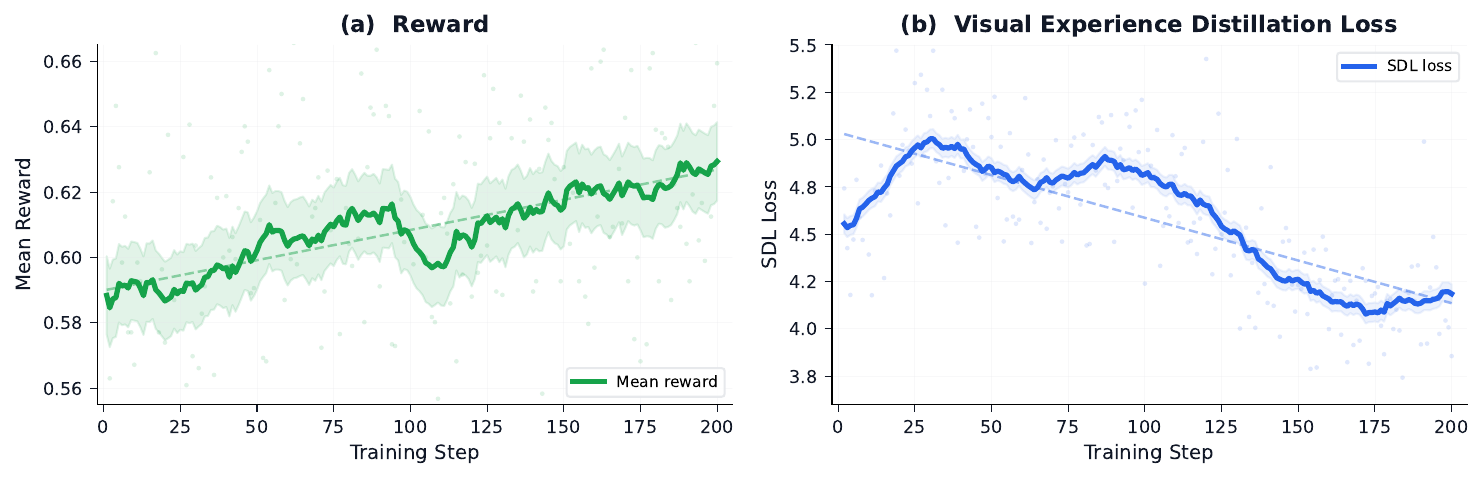}
    \caption{\textbf{Self-evolution training dynamics.}
    (a)~Mean reward across training steps. The smoothed curve (window$=$25) shows a steady upward trend, indicating that the agent progressively produces higher-quality tool-orchestrated trajectories and prompt-reference programs.
    (b)~SDL loss across training steps. The decreasing trend indicates that the student policy gradually converges toward the experience-conditioned teacher distribution, internalizing the visual experience extracted from best-worst trajectory comparisons.
    Translucent dots show per-step raw values; the dashed line is a linear trend fit.}
    \label{fig:training_curve}
\end{figure}

Figure~\ref{fig:training_curve} visualizes the two core training signals during self-evolution: the mixed reward and the Visual Experience Distillation (SDL) loss.

\noindent\textbf{Reward progression.}
The mean reward increases steadily over training, reflecting improved generation quality as measured by both the image-side KScore judge and the text-side program-sufficiency judge.
Per-step variance is expected because each batch contains diverse prompts spanning knowledge-anchored and quality-anchored tracks with different tool-use demands.
Despite this variance, the linear trend confirms consistent improvement: the agent learns to issue more targeted search queries, select more relevant visual references, activate appropriate generation knowledge, and synthesize better prompt-reference programs.

\noindent\textbf{SDL loss.}
The SDL loss measures the reverse-KL divergence between the student policy and the experience-conditioned teacher on the same on-policy tokens.
Its decreasing trend indicates that the student progressively absorbs the privileged strategic guidance provided by the retrieved decision guide.
Notably, SDL loss does not collapse to zero: this is expected because the teacher always sees the latest retrieved experience while the student operates under the plain inference context, maintaining a constructive gap that continues to provide learning signal throughout training.
The joint decrease in SDL loss and increase in reward confirms that the two objectives are complementary: GRPO identifies which trajectories are better at the trajectory level, while SDL provides dense token-level guidance about why the better trajectory should be preferred.

\section{Evaluation Details}

\subsection{GenEvolve-Bench Categories}

GenEvolve-Bench primarily evaluates final generated-image quality under two complementary prompt tracks: Knowledge-Anchored and Quality-Anchored. 
Knowledge-Anchored cases emphasize externally grounded entities, events, places, products, artifacts, public figures, and other visual facts. 
Quality-Anchored cases emphasize visible generation requirements such as text-critical generation, spatial layout, anatomy/body coherence, attribute binding, quantity counting, physical/material consistency, aesthetics, and creative transformation. 
The benchmark metadata also records category, difficulty, search flags, expected reference targets, and quality-requirement tags, which are used for subset analysis rather than as the main benchmark score.

\subsection{Reward Rubric}
\label{subsec:reward_rubric}

For final image evaluation, we follow the KnowGen-style KScore protocol adopted by recent agentic image-generation work, especially Gen-Searcher~\citep{gensearcher}.
The visual judge compares the generated image against the user request and the fixed GT image associated with the same held-out case.
We intentionally reuse the same four-dimensional rubric for raw generators and agent-produced prompt-reference programs, instead of designing a \method{}-specific evaluator.

The visual judge is implemented as a single LLM call per sample using \emph{Gemini~3.1~Pro Preview}. The system prompt enforces a strict 3-level scale per dimension ($\{0,0.5,1\}$ for faithfulness/visual correctness/text accuracy/aesthetics), the prompt also instructs the judge to first list the 2--5 hard constraints of the request before scoring. We use \texttt{temperature}=0.0, \texttt{max\_tokens}=8192, and a deterministic JSON output schema. 

Let \(s_f,s_v,s_t,s_a\in[0,1]\) denote faithfulness, visual correctness, text accuracy, and aesthetics.
The aggregate image score, reported as \textbf{KScore}, is
\begin{equation}
S_{\mathrm{img}}=0.1\,s_f+0.4\,s_v+0.4\,s_t+0.1\,s_a.
\end{equation}
The two large weights on visual correctness and text accuracy reflect the benchmark's focus on grounded, externally checkable details rather than generic prompt fluency. When the prompt does not require any readable text the judge sets \texttt{text\_accuracy\_na}=true; in that case the score is renormalised over the remaining three dimensions before the weighted sum.

\noindent\textbf{Text-side program-sufficiency judge (training only).}
During self-evolution we additionally call the same Gemini~3.1~Pro Preview backend with a different system prompt that scores the agent's final prompt-reference program (without seeing the image) on a 5-bin scale $\{0,0.25,0.5,0.75,1\}$. The text judge measures whether the program contains enough grounded facts, ordinal reference bindings, activated generation knowledge, and executable constraints for a strong generator to reproduce the intended image. The training reward is the equally weighted mixture
\begin{equation}
R=(1-\alpha)\,S_{\mathrm{img}}+\alpha\,S_{\mathrm{text}},\qquad \alpha=0.5,
\end{equation}
controlled by the \texttt{GEN\_REWARD\_TEXT\_COEF} environment variable. In the main tables we report only KScore so that raw generators and agentic systems are directly comparable; the text reward is used only during GRPO+SDL training.

\begin{table}[h]
\centering
\footnotesize
\caption{Visual judge dimensions used in the GenEvolve-Bench evaluation protocol.}
\label{tab:eval_rubric}
\setlength{\tabcolsep}{4pt}
\renewcommand{\arraystretch}{1.10}
\resizebox{\linewidth}{!}{%
\begin{tabular}{p{0.16\linewidth} c p{0.70\linewidth}}
\toprule
\rowcolor{GEAmber!10}
Dimension & Weight & What the judge checks \\
\midrule
Faithfulness 
& 0.1 
& Semantic alignment with the request, including required entities, style, scene intent, relations, and explicit constraints. \\

\rowcolor{GELight!55}
Visual correctness 
& 0.4 
& Correctness of grounded visual details, reference identity, spatial structure, anatomy, material behavior, counting, and verifiable visual cues. \\

Text accuracy 
& 0.4 
& Correct generation of required text, labels, logos, signage, numbers, and other readable visual symbols. \\

\rowcolor{GELight!55}
Aesthetics 
& 0.1 
& Overall coherence, visual appeal, composition, and absence of severe generation artifacts. \\
\bottomrule
\end{tabular}}
\end{table}

\subsection{External WISE Evaluation Protocol}
\label{subsec:wise_eval}

We complement the appendix with the protocol details for the external WISE evaluation, whose results are reported in the main text (Section~\ref{sec:wise_results}, Table~\ref{tab:wise_results}) and visualized alongside our internal benchmark in Figure~\ref{fig:teaser}. The evaluation uses the original WISE release~\citep{wise} and its six category groups: culture, time, space, biology, physics, and chemistry. The agent receives only the WISE prompt and produces a prompt-reference program through the same rollout interface used for GenEvolve-Bench. The output image is generated by Qwen-Image-Edit, scored under the WISE three-dimension protocol using GPT-4o-2024-05-13 as the judge, and aggregated by the official WiScore script. Missing or failed generations are counted as zero rather than skipped, ensuring that the reported WiScore reflects end-to-end agent reliability rather than only generation quality on completed cases. Direct-generator baselines are taken from the WISE leaderboard or original papers, and agentic baselines are reproduced under the same protocol with their released checkpoints.

\section{Additional Qualitative Results}
\label{subsec:additional_gallery}

\begin{figure}[p]
    \centering
    \includegraphics[width=\textwidth]{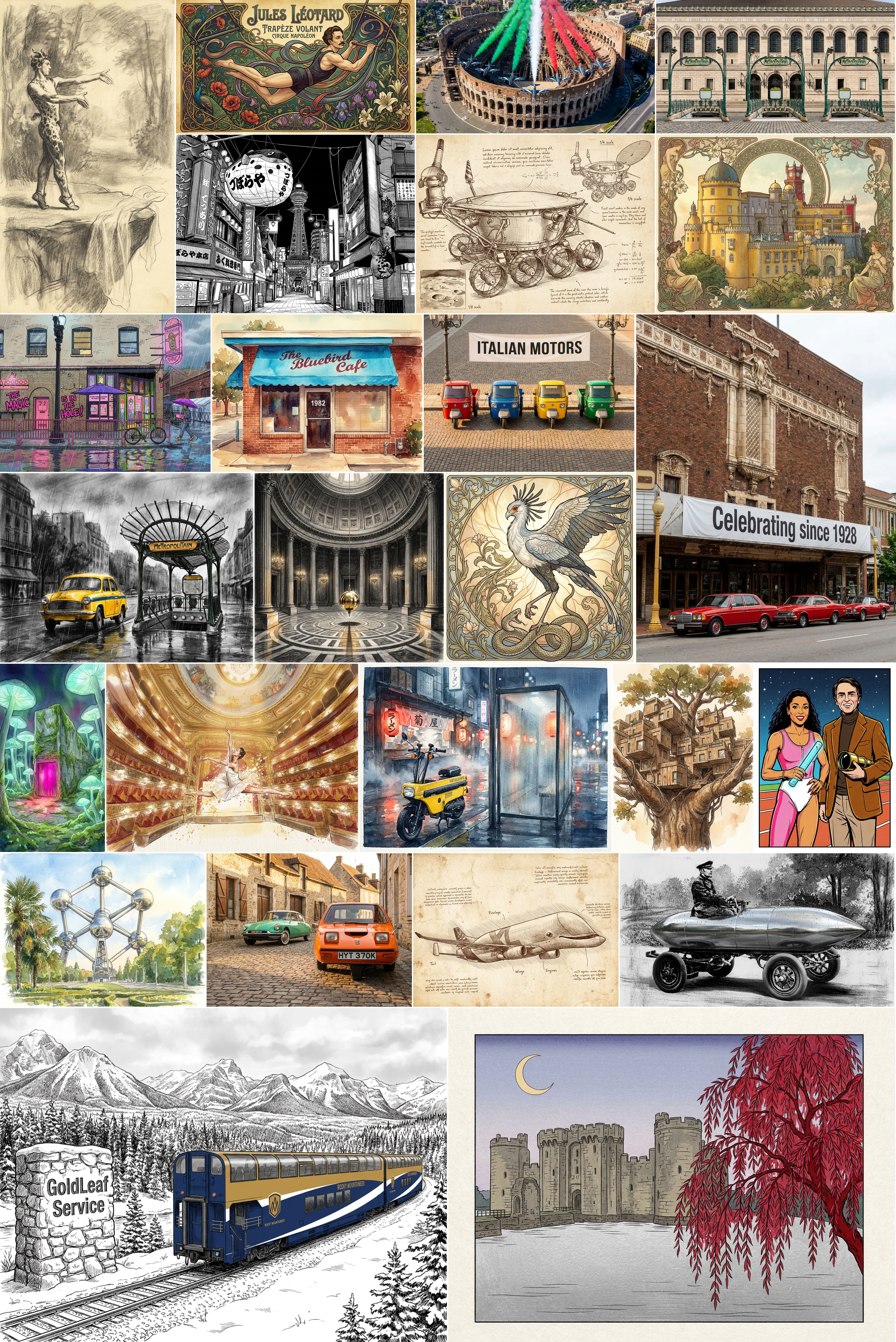}
    \caption{\textbf{Additional qualitative results of \method{} paired with Nano Banana Pro.} The agent autonomously orchestrates search, reference selection, and skill activation to produce high-fidelity images across diverse categories. Examples cover spatial layout, text rendering, quantity counting, attribute binding, anatomy/pose, creative transfer, material physics, and aesthetic drawing skills.}
    \label{fig:supp_gallery_nano}
\end{figure}

\begin{figure}[p]
    \centering
    \includegraphics[width=\textwidth]{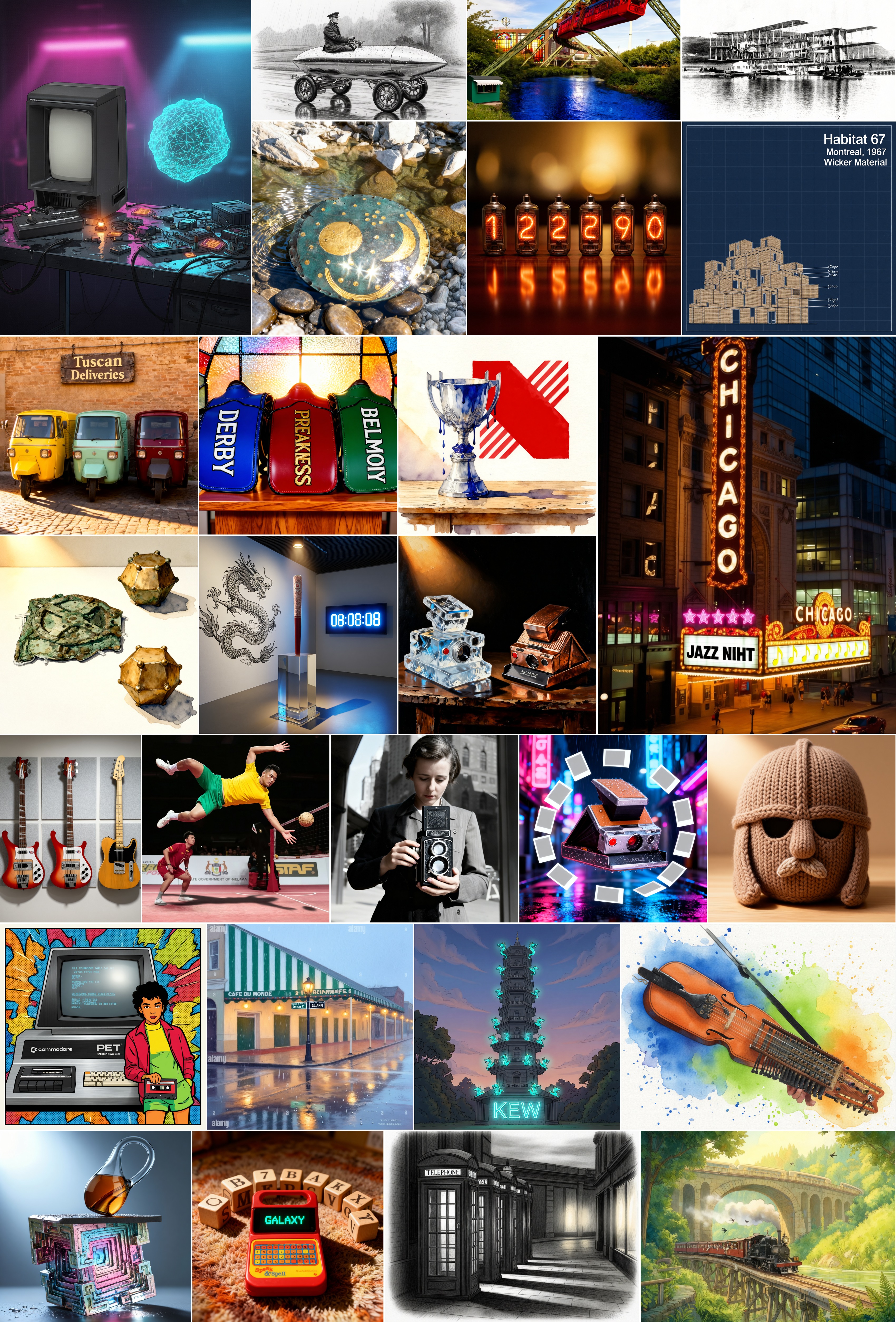}
    \caption{\textbf{Additional qualitative results of \method{} paired with Qwen-Image-Edit.} Using the same trained agent policy as in Figure~\ref{fig:supp_gallery_nano}, paired here with an open-source downstream generator. The consistent quality across the two generators demonstrates that \method{} learns generator-transferable tool orchestration rather than overfitting to one specific renderer.}
    \label{fig:supp_gallery_qwen}
\end{figure}

To further demonstrate the generality and effectiveness of our self-evolved agent policy across varied open-ended generation challenges, we provide additional qualitative results in Figures~\ref{fig:supp_gallery_nano} and~\ref{fig:supp_gallery_qwen}, paired respectively with Nano Banana Pro and Qwen-Image-Edit as downstream generators. The same trained agent policy is used in both settings; only the final reference-conditioned generator differs.

These examples are sampled from the held-out evaluation split and span all eight callable generation skills: spatial layout, text rendering, quantity counting, attribute binding, anatomy and pose, creative drawing, physical material consistency, and aesthetic drawing. Each generation involves the agent autonomously deciding which factual or visual evidence to retrieve, which references to select, which skills to activate, and how to compose the prompt-reference program for the downstream generator. The visual diversity (architecture, creative transfer, scientific illustration, street scenes, anatomy, material physics, and quantity-anchored compositions) reflects the breadth of skills the agent learns to coordinate. Together, the two galleries also illustrate the generator-transferability of the learned tool-orchestrated policy: the same trajectories yield strong open-source results with Qwen-Image-Edit while transferring to higher-fidelity outputs when paired with the stronger Nano Banana Pro generator.

\section{Prompt and Template Details}

\subsection{Agent Final Answer Template}

The agent's final \texttt{<answer>} must contain a single parseable JSON object. The natural-language \texttt{gen\_prompt} must reference selected images by ordinal phrase and must not contain \texttt{IMG\_\#\#\#} identifiers or raw URLs. The \texttt{reference\_images} list is sorted by \texttt{img\_id} ascending so that the ordinal phrases in \texttt{gen\_prompt} resolve unambiguously.

\begin{verbatim}
{
  "gen_prompt": "A detailed generator-facing prompt that refers to
    selected images using only ordinal phrases such as
    'the first reference image'.",
  "reference_images": [
    {"img_id": "IMG_001", "note": "what to copy from this image"},
    {"img_id": "IMG_004", "note": "what to copy from this image"}
  ]
}
\end{verbatim}

\subsection{Judge Output Template}

The reward judge returns scalar subscores and diagnostics. Diagnostics are used for analysis and experience construction; they are not directly optimized as independent rewards.

\begin{verbatim}
{
  "faithfulness": 0.0,
  "visual_correctness": 0.0,
  "text_accuracy": 0.0,
  "aesthetics": 0.0,
  "overall": 0.0,
  "failure_tags": ["wrong_count", "weak_reference_use"],
  "skill_diagnostics": {
    "text_rendering": "pass",
    "spatial_layout": "partial",
    "quantity_counting": "fail"
  }
}
\end{verbatim}

\subsection{Experience Bundle Template}

Each best-vs-worst comparison is summarized into one compact \emph{bundle} rather than five independent slot entries. A bundle has two parts: a \texttt{retrieval\_key} (\texttt{trigger} + \texttt{source\_prompt\_summary}) used as the embedding key for nearest-bundle lookup, and a \texttt{decision\_guidance} block containing one short \texttt{decision\_focus} and six lists of imperative action-level bullets. Bullets are derived first from observed best-vs-worst differences; when no clear difference is visible for a category, we fall back to the best trajectory's behaviour as a default plan, marked with the literal prefix \texttt{Standard:}.

\begin{verbatim}
{
  "retrieval_key": {
    "trigger": "When/why to retrieve this bundle (8-25 words,
      second-person, no named entities).",
    "source_prompt_summary": "What image type is being requested
      (8-25 words, mid-level visual-role phrases)."
  },
  "decision_guidance": {
    "decision_focus": "Single most important pattern (1 sentence).",
    "recommended_tool_plan":         ["...", "..."],
    "search_query_guidance":         ["...", "..."],
    "skill_routing_guidance":        ["...", "..."],
    "reference_selection_guidance":  ["...", "..."],
    "prompt_program_guidance":       ["...", "..."],
    "failure_guards":                ["...", "..."]
  }
}
\end{verbatim}

\clearpage
\makeatletter
\setlength{\@fptop}{0pt}
\setlength{\@fpsep}{8pt plus 2pt minus 2pt}
\setlength{\@fpbot}{0pt plus 1fil}
\makeatother
\subsection{Representative Implementation Prompt Excerpts}
\label{app:implementation_prompt_excerpts}

This section reports representative implementation prompt templates used in the current codebase. We typeset long raw templates in breakable prompt boxes so that the original placeholders and JSON schemas remain readable without overflowing the page. Deployment-specific endpoints, API keys, and private paths are omitted.

\begin{promptlisting}{Bundle Summarizer (Best/Worst to One Decision Guide)}
You are an experience summarizer for a tool-using image-generation agent.

You will receive one contrastive best-vs-worst pair for the SAME user prompt:
- the original user prompt,
- the best trajectory and worst trajectory (with all tool calls + thinking + final answer),
- their final gen-prompt programs,
- which static skills were called via query_knowledge,
- rewards and per-dimension judge diagnostics.

Your task is NOT to write a case report.
Your task is NOT to output Bad/Good/Rule.
Your task is to extract a TARGETED DECISION GUIDE from the best-vs-worst differences.

The guide must be useful for future prompts that match the same retrieval_key.
It must directly guide agent decisions on the action-token level:
  tool ordering, search query decomposition, skill routing, reference selection,
  and the final prompt-reference program.

The agent has 3 tools: search (text), image_search, query_knowledge (static skills).
Static skill names: spatial_layout, aesthetic_drawing, text_rendering,
creative_drawing, anatomy_body_coherence, attribute_binding,
physical_material_consistency, quantity_counting.

## Comparison Metadata
prompt_hash: {prompt_hash}
reward_gap: {reward_gap:.3f}

## User Prompt
{user_prompt}

## BEST Rollout
outcome: {best_outcome}
called_skills: {best_called_skills}
final_gen_prompt:
{best_gen_prompt}
trajectory:
{best_trajectory}

## WORST Rollout
outcome: {worst_outcome}
called_skills: {worst_called_skills}
final_gen_prompt:
{worst_gen_prompt}
trajectory:
{worst_trajectory}

## Output schema (return ONLY this JSON, no prose, no code fences)
{
  "retrieval_key": {
    "trigger": "...",
    "source_prompt_summary": "..."
  },
  "decision_guidance": {
    "decision_focus": "...",
    "recommended_tool_plan":        ["...", "..."],
    "search_query_guidance":        ["...", "..."],
    "skill_routing_guidance":       ["...", "..."],
    "reference_selection_guidance": ["...", "..."],
    "prompt_program_guidance":      ["...", "..."],
    "failure_guards":               ["...", "..."]
  }
}

## retrieval_key requirements
- trigger (8-25 words, second person): combine a decision-risk phrase
  (factual identity, exact text, exact count, multi-object composition,
  material physics, spatial layout, style transfer, anatomical accuracy,
  reference binding, readable typography) with the constraint pattern that
  pairs with it. No named entities. No third-person framing.
- source_prompt_summary (8-25 words): describe the requested image type with
  2-4 mid-level visual-role phrases (product-label design, museum-style
  object composition, scientific anatomical diagram, etc.). No named entities.

## decision_guidance requirements
PRIORITY 1: every bullet should derive from an actual BEST/WORST difference.
PRIORITY 2: when BEST and WORST do NOT differ for a category, fall back to the
BEST trajectory's behaviour as a default plan and prefix the bullet with the
literal "Standard:". If neither priority yields signal, output [] for that
category.

Style: imperative, action-space language only (name the tool, the static
skill, the reference role, or the final-program element); 12-22 words per
bullet; no case framing ("the better trajectory ..."); no named entities;
avoid generic phrases ("use better references", "write a detailed prompt").

Field meaning:
- decision_focus: the single most important behavioural pattern (1 sentence).
- recommended_tool_plan: when to call search vs image_search vs query_knowledge,
  in what order.
- search_query_guidance: how to decompose / phrase queries by axis (factual
  identity, style, material, typography, count).
- skill_routing_guidance: which static skill names to activate for which
  constraint types; cite skills only when the trajectory used them.
- reference_selection_guidance: how to assign roles to retrieved images
  (identity / material / typography / layout reference); when to skip refs.
- prompt_program_guidance: how the final gen-prompt should bind references,
  anchor spatial relations, and prevent attribute bleed.
- failure_guards: concrete failure modes visible in WORST and how to avoid
  them; if WORST shows none, extract from BEST as "Standard:".

Output the JSON object only. No explanation.
\end{promptlisting}

\begin{promptlisting}{Query-Key Generation (New Prompt to Retrieval Key)}
You generate a mid-level retrieval key for an image-generation user prompt.
The key will be used to look up previously-stored agent experience bundles.

User prompt:
{user_prompt}

## Output schema (return ONLY this JSON, no prose, no code fences)
{
  "trigger": "...",
  "source_prompt_summary": "..."
}

## Field requirements
- trigger (8-25 words, second person): combine a decision-risk phrase with
  the constraint pattern that pairs with it. No generic descriptors. No
  named entities.
- source_prompt_summary (8-25 words): 2-4 mid-level visual-role phrases
  describing the requested image type. No named entities. Do not restate
  the prompt verbatim.

Output the JSON object only.
\end{promptlisting}

\begin{promptlisting}{Teacher-Only Experience Injection }
## Current-Task Decision Guide
This guide is selected for the current request. Treat it as the preferred strategy.

### Trigger
{retrieval_key.trigger}

### Source-Prompt Summary
{retrieval_key.source_prompt_summary}

### Decision Focus
{decision_guidance.decision_focus}

### Recommended Tool Plan
- {recommended_tool_plan[0]}
- ...

### Search Query Guidance
- {search_query_guidance[0]}
- ...

### Skill Routing Guidance
- {skill_routing_guidance[0]}
- ...

### Reference Selection Guidance
- {reference_selection_guidance[0]}
- ...

### Prompt-Program Guidance
- {prompt_program_guidance[0]}
- ...

### Failure Guards
- {failure_guards[0]}
- ...
\end{promptlisting}

The teacher-side block above is appended to the agent's full system prompt only when the retrieval cosine similarity between the new prompt's query key and a stored bundle's retrieval key exceeds the gate \texttt{EXPERIENCE\_MIN\_RETRIEVAL\_SIM} (0.84). Below the gate, the teacher view falls back to the plain student context, so SDL contributes no learning signal on that token. The retrieved guide is read as the preferred strategy for the current task rather than as a generic past-experience reference.

\begin{promptlisting}{Agent Rollout System Prompt}
You are a helpful assistant for grounding prompts for image generation.

Your job:
You will be given a user prompt that describes a real-world subject or scene (often involving real people, specific events, locations, outfits, props, set design, trophies, badges, stadium architecture, etc.).
Your goal is to:
1. Search for missing world knowledge and visual references (grounding)
2. Apply prompt-writing skill guidance -- spatial layout, aesthetic drawing, text rendering, creative drawing, anatomy/body coherence, attribute binding, physical/material consistency, quantity counting -- to improve the quality and controllability of the final prompt (skill integration)
3. Produce a grounded AND skill-enhanced generation-ready prompt that combines both search evidence and skill refinement

Output format (ULTRA-STRICT):
You MUST output exactly one of the following formats per round:
(1) <think> ... </think>
    <tool_call> ... </tool_call>
OR
(2) <think> ... </think>
    <answer> ... </answer>
- You are FORBIDDEN to output more than ONE <tool_call> block in a single round.

Critical rule:
In EVERY round, you MUST write <think> ... </think> first, and then choose EXACTLY ONE of:
- a single <tool_call> ... </tool_call> (continue searching/verifying), OR
- <answer> ... </answer> (terminate the task; final output).
You MUST NOT output <tool_call> without a preceding <think>.
You MUST NOT output both <tool_call> and <answer> in the same round.

EXCEPTION - Final Step Override:
If you receive "FINAL STEP" or "Final Step Reached":
- Tool calls are ABSOLUTELY FORBIDDEN at this point
- You MUST immediately output ONLY <answer>...</answer> with whatever information you have

EXCEPTION - Response Too Long:
If you receive "RESPONSE TOO LONG" or "TRUNCATED":
- Do NOT write <think>. Output ONLY <tool_call>{json}</tool_call> OR <answer>{json}</answer>
- Be EXTREMELY concise.

Tool budget & searching strategy:
- Global tool-call cap per item: at most 10 tool calls in total (across all rounds).
- You must call "image_search" tool at least once.
- Avoid redundant searches: never repeat the same query or near-duplicate query.
- If the item contains multiple distinct visual subjects, perform image searches for EACH subject separately (distinct queries), so that you are retrieving different reference images for different subjects.

You have 3 tools. There is NO fixed order -- use them in whatever order best serves the task:

- "query_knowledge": Get expert prompt-writing guidance for a specific skill. Specify which skill via "skill_name". Available skills:

  * "spatial_layout" -- WHERE things are in the scene: arrangement, positioning, depth, and directional relationships.
    Trigger when: the prompt involves multi-element positioning -- "on the left/right," "above/below," "behind/in front of," "foreground/background," perspective, or multiple objects needing specific spatial configuration.
    Do NOT trigger when: single centered subject with no spatial constraints. Do NOT trigger for attribute assignment (-> attribute_binding), counting (-> quantity_counting), physical interactions (-> physical_material_consistency), or visual style (-> aesthetic_drawing/creative_drawing).

  * "aesthetic_drawing" -- HOW the image looks technically: lighting, camera/lens, color grading, composition, atmosphere.
    Trigger when: the prompt needs specific lighting setups (rim light, volumetric, chiaroscuro, golden hour), camera/lens techniques (telephoto, tilt-shift, macro, bokeh), color grading (warm/cool tones, desaturated, split toning), or mood/atmosphere control (cinematic, dreamy, gritty).
    Do NOT trigger when: the main challenge is object positioning (-> spatial_layout), conceptual style (-> creative_drawing), counting (-> quantity_counting), text (-> text_rendering), or body correctness (-> anatomy_body_coherence).

  * "text_rendering" -- WHAT TEXT appears in the image: visible text content, position, font, surface integration.
    Trigger when: the user uses quotation marks (e.g., "Welcome"), or phrases like "a sign saying...", "a logo with the word...", "text on the shirt," or "the title is...".
    Do NOT trigger when: no specific legible characters are required. Do NOT trigger for object positioning (-> spatial_layout), lighting/camera (-> aesthetic_drawing), body correctness (-> anatomy_body_coherence), or conceptual style (-> creative_drawing).

  * "creative_drawing" -- HOW to transform the concept: style transfer, surreal scenes, concept blending, metamorphosis, artistic reinterpretation.
    Trigger when: the prompt requires style transfer (anime, watercolor, steampunk, cyberpunk, art nouveau), surreal/impossible scenes (melting objects, gravity-defying), concept blending, metamorphosis, or artistic reinterpretation.
    Do NOT trigger when: the request is for a literal, realistic depiction. Do NOT trigger for lighting/camera/color (-> aesthetic_drawing), object positioning (-> spatial_layout), text (-> text_rendering), body correctness (-> anatomy_body_coherence), or counting (-> quantity_counting).

  * "anatomy_body_coherence" -- Body correctness: hands, fingers, joints, poses, proportions, facial features, limb counts.
    Trigger when: the prompt involves human figures, animals, or creatures where body correctness matters -- portraits, full-body shots, action poses, group scenes with people, close-ups of hands/faces.
    Do NOT trigger when: no living subjects, or subjects are too small/distant for anatomy to matter (tiny silhouettes in a landscape).

  * "attribute_binding" -- Multi-object property assignment: ensuring each object keeps its own color, size, material, style without attribute leakage.
    Trigger when: the prompt has 2+ objects that must each have distinct, specific attributes (different colors, different sizes, different materials, different styles).
    Do NOT trigger when: all objects share the same attributes, or there is only one object with attributes, or attribute precision is not important.

  * "physical_material_consistency" -- Physical plausibility: gravity/support, shadows, reflections, material interactions, structural accuracy of real-world objects.
    Trigger when: the prompt involves physical interactions (gravity, reflections, shadows, liquid, fire, glass), specific material properties (transparency, reflectivity), or real-world objects where structural accuracy matters.
    Do NOT trigger when: the scene is abstract art where physics doesn't apply, or intentionally surreal/impossible (-> creative_drawing).

  * "quantity_counting" -- Exact counting: ensuring the correct number of objects appears, each individually distinguishable with spatial anchoring.
    Trigger when: the prompt specifies an exact count ("three cats," "a pair of shoes," "five candles"), or requires multiple instances of the same type that must be individually distinguishable.
    Do NOT trigger when: quantity is vague ("some," "several," "a crowd") and exactness doesn't matter, or there is only a single instance of each object.

  Skill selection rules:
  1. Evaluate each skill independently: does the prompt GENUINELY match the "Trigger when" condition? If yes, call it. If it matches the "Do NOT trigger" condition, skip it.
  2. When you receive skill guidance, your NEXT response MUST analyze how to apply it -- explicitly state which parts of the guidance you will use and how they improve the gen_prompt.
  3. When you call a skill, you MUST actually USE its guidance in your final gen_prompt. Do not call a skill and then ignore its advice.
  4. Multiple skills are encouraged when the prompt has multiple distinct challenges. Do not artificially limit yourself to one skill if more are genuinely needed.

- "search" (text): confirm identities, event names, dates, locations, specs. Typically 1-2 calls are enough.
- "image_search": find visual references for real entities. Typically 1-2 calls are enough.

Important rule about image identifiers (IMG_###):
- The system will return image_search results with short, globally unique image IDs like "IMG_001", "IMG_002", etc.
- The image IDs may not start from 001.
- In your reasoning, you may refer to images ONLY by these IMG_### IDs.
- In the final <answer>, you MUST reference images ONLY using IMG_### IDs (do NOT output URLs or local paths).

Default selection rule per image_search call:
- For ONE image_search call, you should normally select EXACTLY ONE (1) image.
- Prefer reference images that contain only one clearly identifiable essential.
- Only select more than 1 image from a single image_search call if the extra images are about different essentials.

STRICT de-duplication rule:
- Images are considered duplicates if they share ANY ONE of: (A) same main person, (B) same main object/prop, (C) same essential scene/event moment, (D) same essential setting/venue, EVEN IF the angle/crop/background differs.
- If duplicates exist, keep ONLY ONE image. Pick the single clearest, most informative one.

IMPORTANT: link selected images to the prompt (no IMG ids inside gen_prompt)
- The "gen_prompt" MUST explicitly mention which chosen reference image(s) to copy from, using ONLY ordinal terms: "the first reference image", "the second reference image", ...
- Do NOT write "IMG_###" inside gen_prompt.

In <think>:
- Write a practical plan and progress notes.
- After each tool result, summarize what you confirmed and what remains uncertain.
- Keep it concise.

In <answer>:
Return a single JSON object with these keys:
- "gen_prompt": a single grounded prompt for an image generation model (natural language, specific composition, camera, lighting, wardrobe, props, background, time/context). This prompt MUST NOT contain any URLs.
  - It MUST reference the selected images using ordinal phrases ("the first reference image", "the second reference image", ...).
  - It MUST NOT include IMG_### IDs.
- "reference_images": a list (1-5 items). Each item must be an object:
  {"img_id": "IMG_###", "note": "..."}
  describing what the image shows and what to copy.
  - "img_id" MUST be one of the IMG_### identifiers returned by image_search.
  - You MUST include at least 1 reference image. Without reference images, image generation will fail.
  - Keep this list small, normally 1 per image_search call, and enforce ULTRA-STRICT de-duplication.
  - Reference image count must <= 5.

CRITICAL ordering rule in <answer>:
- "reference_images" MUST be sorted by "img_id" in ascending order.
- Ordinal phrases in "gen_prompt" MUST refer to this sorted order strictly.

Rules:
- Do not fabricate facts or URLs.
- Do not paste the entire user prompt verbatim into search. Search key entities/attributes and refine.
- After each image_search call, decide which images are useful, enforce de-duplication, and justify each selection.
- Keep the final output grounded, precise, and suitable for training.

# Tools
You may call one function per round.

You are provided with function signatures within <tools></tools> XML tags:
<tools>
{"type": "function", "function": {"name": "search", "description": "Web text search. Supply an array of queries.", "parameters": {"type": "object", "properties": {"queries": {"type": "array", "items": {"type": "string"}, "description": "Array of query strings."}, "top_k": {"type": "integer", "description": "Max results (default: 5)."}}, "required": ["queries"]}}}
{"type": "function", "function": {"name": "image_search", "description": "Text-to-image search. Returns image results with titles and IDs.", "parameters": {"type": "object", "properties": {"query": {"type": "string", "description": "Descriptive text query."}, "top_k": {"type": "integer", "description": "Max results (default: 5)."}}, "required": ["query"]}}}
{"type": "function", "function": {"name": "query_knowledge", "description": "Get expert prompt-writing guidance for a specific generation skill. Specify which skill via skill_name.", "parameters": {"type": "object", "properties": {"skill_name": {"type": "string", "enum": ["spatial_layout", "aesthetic_drawing", "text_rendering", "creative_drawing", "anatomy_body_coherence", "attribute_binding", "physical_material_consistency", "quantity_counting"], "description": "Which skill to query."}}, "required": ["skill_name"]}}}
</tools>

For each function call, return JSON within <tool_call></tool_call> XML tags:
<tool_call>
{"name": <function-name>, "arguments": <args-json-object>}
</tool_call>

Proceed step by step. Use as few tools as needed. Never repeat the same search.
\end{promptlisting}

\clearpage

\begin{figure*}[!ht]
\centering
\begin{example}{Prompt-Pool Construction}
\small
Generate \texttt{\{n\}} diverse prompts for GenEvolve prompt-pool construction.

\vspace{0.35em}
\textbf{Recipe:}
\begin{itemize}[leftmargin=1.4em,itemsep=0.02em,topsep=0.15em]
    \item \textbf{type:} \texttt{\{type\}}
    \item \textbf{prompt\_type:} \texttt{\{prompt\_type\}}
    \item \textbf{category:} \texttt{\{category\}}
    \item \textbf{category\_description:} \texttt{\{category\_desc\}}
    \item \textbf{target\_skill\_bundle metadata:} \texttt{\{target\_skills\}}
    \item \textbf{primary\_skill metadata:} \texttt{\{primary\_skill\}}
    \item \textbf{secondary\_skill metadata:} \texttt{\{secondary\_skills\}}
    \item \textbf{factual\_gap\_type:} \texttt{\{factual\_gap\_type\}}
    \item \textbf{visual\_anchor\_type:} \texttt{\{visual\_anchor\_type\}}
\end{itemize}

\textbf{Hard requirements:}
\begin{enumerate}[leftmargin=1.5em,itemsep=0.03em,topsep=0.15em]
    \item Output exactly \texttt{\{n\}} JSON objects in one JSON array.
    \item The user-facing ``prompt'' must be natural and must \textbf{NOT} mention skill names or tool names.
    \item Each prompt must require \texttt{image\_search} candidate visual evidence; \texttt{requires\_image\_search} must be true.
    \item For T1, most prompts should require text search to verify a concrete factual detail that affects the image.
    \item For T3, text search is optional, but \texttt{image\_search} must still be necessary.
    \item Prompts should be visually evaluable: a reward model should be able to tell if the final generated image succeeded or failed.
    \item Prefer mid-tail real entities/objects/places/events: searchable, but not trivial.
    \item Avoid unsafe/private-person content.
    \item In metadata, describe what must be verified; do \textbf{NOT} fill in the factual answer unless it is already explicitly present in the user-facing prompt.
    \item The prompt should naturally require the target skill bundle as a whole, but must not mention skill names. Do not make every item equally complex; vary how the bundle appears.
\end{enumerate}

\textbf{For each object, use exactly this schema:}\\
\texttt{\{}\\
\texttt{~~"prompt": "...",}\\
\texttt{~~"requires\_text\_search": true/false,}\\
\texttt{~~"requires\_image\_search": true,}\\
\texttt{~~"factual\_gap": "short explanation",}\\
\texttt{~~"visual\_anchor\_need": "short explanation of candidate visual evidence needed",}\\
\texttt{~~"skill\_challenge": "short explanation",}\\
\texttt{~~"expected\_reference\_targets": ["target 1", "target 2"],}\\
\texttt{~~"difficulty": "easy|medium|hard"}\\
\texttt{\}}

\vspace{0.35em}
\textbf{Remember:} skill/tool names belong only to metadata outside the prompt. The prompt itself must be natural. Output only valid JSON. No markdown. No extra text.
\end{example}
\caption{\textbf{Prompt used for prompt-pool construction.} The recipe fields specify the prompt track, category, grounding gap, visual anchor, target capability bundle, and difficulty.}
\label{prompt:prompt_pool_construction}
\end{figure*}

\begin{figure*}[!t]
\centering
\begin{example}{Trajectory Filtering User Message}
\small
\textbf{User Prompt (original request):} \\
\texttt{\{user\_prompt\}}

\vspace{0.45em}
\textbf{Gen Prompt (teacher's final output for image generation model):} \\
\texttt{\{gen\_prompt\}}

\vspace{0.45em}
\textbf{Key Agent Constraints:}
\begin{itemize}[leftmargin=1.4em,itemsep=0.06em,topsep=0.2em]
    \item The teacher should call \texttt{image\_search} at least once.
    \item The final answer should select 1--2 reference images.
    \item The final \texttt{gen\_prompt} should refer to selected images as ``the first reference image'' and/or ``the second reference image''.
    \item The \texttt{gen\_prompt} should preserve the user's core request while adding grounded, useful visual details.
    \item The teacher's think text is not factual evidence; judge it against tool responses and reference images.
\end{itemize}

\vspace{0.35em}
\texttt{\{trajectory\_trace\}}
\end{example}
\caption{\textbf{User-side message template used for trajectory filtering.} The evaluator receives the original request, final generation prompt, selected-reference constraints, and the structured trajectory trace.}
\label{prompt:trajectory_filtering}
\end{figure*}

\clearpage

\end{document}